\documentclass[11pt, a4paper]{berkeley-paul}

\usepackage[utf8]{inputenc} %
\usepackage[T1]{fontenc}    %
\usepackage{hyperref}       %
\usepackage{url}            %
\usepackage{booktabs}       %
\usepackage{amsfonts}       %
\usepackage{nicefrac}       %
\usepackage{microtype}      %
\usepackage{xcolor}         %
\usepackage{natbib}
\usepackage[capitalise]{cleveref}
\usepackage{graphicx}
\usepackage{algorithm}
\usepackage{caption}
\usepackage{algpseudocode}
\usepackage{wrapfig}
\usepackage{amsmath}
\usepackage{tabularx}
\DeclareMathOperator*{\argmax}{arg\,max}

\title{Test-Time Gradient Guidance of Flow Policies in Reinforcement Learning}

\author{%
  Zhiyuan Zhou$^1$,
  Andy Peng$^1$,
  Charles Xu$^1$,
  Qiyang Li$^1$,
  Jost Tobias Springenberg$^2$, \\
  \Authfont
  Kevin Frans$^1$,
  Sergey Levine$^{1,2}$ \\
  $^1$ UC Berkeley, $^2$ Physical Intelligence
}

\newcommand{\methodname}[0]{QGF}
\newcommand{\methodexplain}{Q-Guided Flow}
\newcommand{\ie}{\emph{i.e.},\ }

\definecolor{themecolor}{HTML}{5C788F}
\definecolor{ourblue}{HTML}{73A4C6}
\hypersetup{colorlinks=true, allcolors=ourblue}

\begin{document}

\runningtitle{Test-Time Gradient Guidance of Flow Policies in Reinforcement Learning}

\teaserfigure{
    \begin{center}
        \vspace{-1em}
        \includegraphics[width=\textwidth]{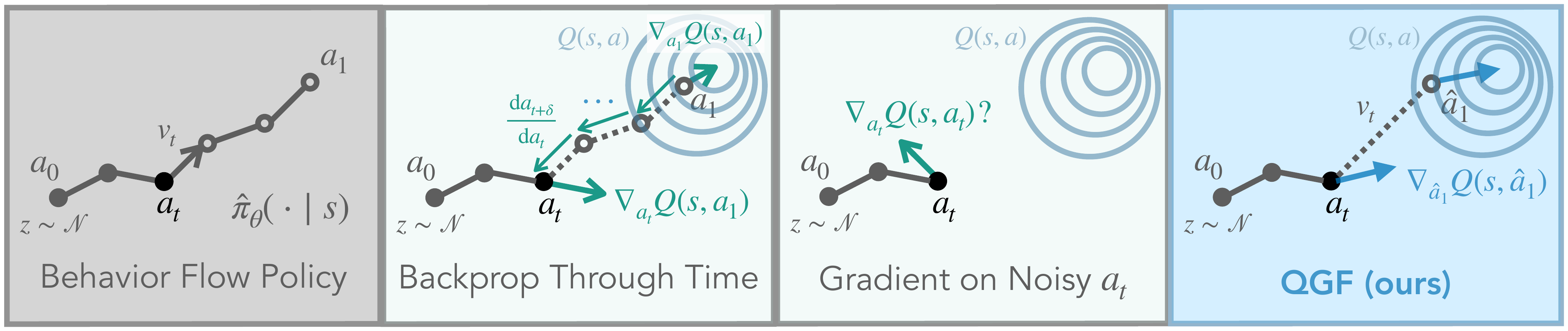}
        \captionof{figure}{We propose \methodname{} (\methodexplain{}), an RL algorithm that guides denoising steps of a policy trained via flow matching with critic gradient at test time. Our critic gradient estimator avoids both taking gradient at noisy action not seen during training and performing expensive, high-variance backpropagation through time, and performs better than other test-time RL methods while being competitive with the best train-time baseline.}
        \label{fig:teaser}
    \end{center}    %
}

\begin{abstract}
Expressive continuous control policies, such as diffusion and flow models, form the backbone of recent advances in scaling imitation learning for simulated and real robot control. While they are known to scale stably in the supervised imitation learning setting, incorporating them into reinforcement learning (RL) pipelines for policy improvement has proven more difficult. It often requires specialized training objectives or backpropagating through denoising processes, which cause well-known issues with stability and affect scalability. In this paper we study the question of whether simple policy improvement schemes at test time alone, leaving stable supervised policy training intact, can be a competitive alternative which sidesteps these issues. To this end, we propose \methodname{} (\methodexplain), an RL algorithm that performs policy optimization entirely at test time. \methodname{} works by pre-training both a reference flow policy (via a standard behavioral cloning objective) and a value function critic and, at test time, using the value gradient to guide the reference policy to generate higher-value actions without any additional policy learning. Empirically, \methodname{} outperforms prior test-time RL methods on single-task and goal-conditioned offline RL benchmarks with high-dimensional action spaces, and is competitive with state-of-the-art training-time algorithms while being much cheaper to run. Moreover, it exhibits favorable scaling with model size by avoiding the instability of actor-critic training, offering a practical and effective alternative RL algorithm with expressive policies.
Code: \href{https://github.com/zhouzypaul/qgf}{ github.com/zhouzypaul/qgf}
\end{abstract}

\maketitle

\section{Introduction}

Reinforcement learning (RL) is a powerful framework for optimizing reward-maximizing behavior, but scaling RL algorithms---especially in the offline or off-policy setting---remains a significant challenge. A key difficulty lies in the complexity and instability of policy optimization: unlike simple supervised learning objectives that optimize towards a known fixed target, RL typically alternates between learning a value function and optimizing an actor to maximize this learned function. This coupled training dynamic makes RL policy training sensitive to hyperparameters and prone to optimization instability~\citep{tsitsiklis1996analysis, fujimoto2018addressing, henderson2018deep}, making it hard to scale. This challenge is particularly acute when using expressive policy classes such as diffusion and flow models. These models are attractive because they can represent complex, multimodal action distributions and have been shown to scale effectively when performing standard supervised learning via behavioral cloning (BC). However, incorporating them into RL pipelines typically requires designing specialized training objectives~\citep{chen2024diffusion, park2025flow, koirala2025flow, chen2025one} or backpropagating through long denoising processes~\citep{wang2022diffusion, he2023diffcps, ding2023consistency, zhang2024entropy, wang2024diffusion, espinosa2025scaling, lv2025flow}, which undermines their scalability and stability.

This tension motivates an alternative approach: \textit{rather than designing a more scalable RL objective for policy optimization, can we train the actor with standard supervised learning and use test-time compute to optimize it against a value function?}
Concretely, we want to train a policy with standard behavior cloning (we will refer to this policy as the reference policy in the following), and learn a critic separately using standard temporal-difference (TD) learning methods. Then, at inference time, we want to use the critic to ``guide'' the reference policy towards higher-value actions, without changing the policy parameters. 
Such an approach would then avoid the instability of needing to train an actor to maximize a changing critic, but instead directly leverage well-understood and scalable supervised objectives during policy training, and apply reward optimization dynamically at test time.

A straightforward way to implement this idea is to combine sampling from the reference policy with  optimization of actions against the learned critic.
Previous successful attempts to use test-time compute in this way have mostly relied on a best-of-N (BFN) sampling strategy \citep{stiennon2020learning, hansen2023idql, li2025reinforcement}, where one draws $N$ actions from the policy and then picks the value-maximizing action according to the critic. However, BFN sampling can be prohibitively expensive in high-dimensional action spaces.
One might wonder whether, instead of sampling, we can rely on the critic gradient $\nabla_a Q(s, a)$ to directly steer action generation towards high-value directions.
Unfortunately, gradient-guidance methods are not straightforward to implement for flow and diffusion models, since actions are generated by an iterative denoising process.
The most na\"{\i}ve way to guide this denoising process is to use the gradient of $Q$ with respect to intermediate ``noisy'' actions by backpropagating through the entire denoising process~\citep{wang2022diffusion, he2023diffcps, ding2023consistency, zhang2024entropy, wang2024diffusion, espinosa2025scaling, lv2025flow}, which is computationally expensive, unstable~\citep{park2025flow}, and has high variance (see \cref{fig:gradient-estimator-variance}). %
One cannot simply avoid such backpropagation because the critic has not been trained to handle intermediate noisy actions, and directly using gradients there leads to biased guidance (see \cref{fig:analysis-1d-denoising}).

In this paper, we introduce a method, \methodname{} (\methodexplain), that effectively utilizes critic guidance during sampling from flow policies. \methodname{} circumvents backpropagation through time and critic gradient at untrained noisy actions by using the critic gradient at an approximate action derived from following the learned velocity flow field for a single, large, Euler integration step (see \cref{fig:teaser}).
We conduct extensive empirical analysis in an offline RL setting (both single-task and goal-conditioned) and show that \methodname{} achieves strong performance compared to both other test-time RL algorithms and state-of-the-art RL algorithms that explicitly train policies to achieve high reward values. In addition, we find that \methodname{} scales well as we increase the model size, since we avoid optimizing the policy against an evolving critic during training, and only train with a stable supervised learning loss. Overall, we establish that gradient-based test-time RL is a promising direction for optimizing generative action models that is stable, scalable, and easy to use in practice.

\section{Related Work}
\label{sec:related-work}

\paragraph{Offline RL} studies how to learn a reward-maximizing policy solely from a fixed dataset without any interactions and environment feedback. The key challenge is to extract optimal behavior from a mixed-quality offline dataset, while preventing erroneous extrapolation that deviates too far from dataset behavior. Prior work has extensively studied both value learning methods that prevent erroneous extrapolations~\citep{an2021uncertainty, garg2023extreme, kostrikov2021offline, kumar2020conservative, xu2023offline} and policy extraction techniques~\citep{chen2022offline, fujimoto2021minimalist, hansen2023idql, nair2020awac, peng2019advantage, tarasov2024revisiting, wu2019behavior}. The predominant approach in this setting optimizes the policy to maximize rewards %
during training, and directly samples from such policy at test time. Our approach, on the other hand, does not train the policy to be reward-seeking. In contrast, we train the policy with standard behavioral cloning to imitate the policy that collected the mixed-quality dataset, and perform policy optimization entirely at test time, with guidance from the value function gradient. 

\paragraph{Flow and diffusion policies for RL.}

Previous works have explored different ways of using expressive models such as diffusion and flow models to train the policy for continuous control in RL, including policy gradient~\citep{ren2024diffusion, mcallister2025flow}, importance-weighting~\citep{pfrommer2025reinforcement, fan2025online, ding2024diffusion, zhang2025energy, chen2025one, xue2025advantage, Dern2025Energy}, and actor-critics~\citep{wang2022diffusion, he2023diffcps, ding2023consistency, zhang2024entropy, wang2024diffusion, lv2025flow, chen2024diffusion, park2025flow, koirala2025flow, chen2025one}. Using these iterative denoising methods for actor-critic style algorithms is especially challenging since the actor $\pi_\theta(a \mid s)$ is trained with the critic gradient $\nabla_\theta Q(s, \pi_\theta(a \mid s))$. For single-step Gaussian policies, the actor can leverage the gradient through the reparameterization trick~\citep{haarnoja2018soft}; for flow and diffusion policies however, leveraging the gradient requires either a noise-conditioned critic~\citep{lu2023contrastive, mao2024diffusion, Attarian2026Update} which is hard to estimate accurately, or backpropagating through a multi-step denoising process~\citep{wang2022diffusion, he2023diffcps, ding2023consistency, zhang2024entropy, wang2024diffusion, lv2025flow}, which is expensive and can be unstable in practice~\citep{park2025flow}. Previous works have attempted to avoid unstable backpropagation by (1) distilling the multi-step policy into a single-step policy~\citep{park2025flow, chen2025one}, (2) using a single-step Euler-integration approximation of the multi-step denoising process~\citep{kang2023efficient}, or (3) simply ignoring the denoising process and taking the critic gradient at noisy actions $a_t$ (\ie $\nabla_{a_t} Q(s, a_t)$)~\citep{psenka2023learning, yang2023policy, fang2024diffusion, li2024learning}.
Approach (1) requires training a separate distilled network, which is expensive and can suffer from expressivity issues; approach (2) is much cheaper and doesn't require additional training, but can suffer from approximation error, though we find it to perform better in practice (see \cref{fig:better-optimizer}); approach (3) is the simplest, but leads to biased guidance (see \cref{fig:analysis-1d-denoising}).
Most similar to our work is EDP~\citep{kang2023efficient}, which takes approach (2) above. %
We also use a single-step Euler integration to approximate the denoising process~\citep{bansal2024universal, efron2011tweedie}. 
However, we leverage the critic gradient with respect to the action, while EDP uses the critic gradient with respect to policy parameters.
More importantly, EDP requires training the actor with RL objectives at training time, whereas our method pre-trains a value function and uses its gradient to optimize policy actions at test-time during the action denoising process of a behavior flow policy. 

\paragraph{Test-time methods in RL} often focus on using a value function to refine actions from a reference policy at test time. This can be done by either using the critic to rank action samples and pick the best one~\citep{li2025reinforcement, hansen2023idql}, or hill-climbing on the critic value with additional gradient steps on fully denoised actions~\citep{mark2024policy, xureinforcement}. 
Another line of methods can also control the optimality of the policy at test time not by using the critic, but by training a policy that is conditioned on optimality~\citep{frans2025diffusion, kumar2019reward}.  
Most similar to our work is QFQL~\citep{jang2025q}, which uses the critic gradient to guide the action denoising process. However, QFQL takes the critic gradient at a noisy action step (\ie $\nabla_{a_t} Q(s, a_t)$), where the noisy action may often go out-of-distribution for the critic that has only been trained on fully denoised actions.
In contrast, our method leverages the critic gradient directly on approximations of fully denoised actions, and performs better in practice (see \cref{fig:main-results,fig:dqc-results}).

\section{Preliminaries and Problem Setting}
\label{sec:backgrond}

\paragraph{Markov decision process and Q-learning.}
We consider a Markov Decision 
Process (MDP)~\citep{sutton1998reinforcement} $\mathcal{M} = (\mathcal{S}, \mathcal{A}, P, r, \gamma)$, where $\mathcal{S}$ is the 
state space, $\mathcal{A}$ is the action space, $P(s' | s, a) : \mathcal{S} \times \mathcal{A} 
\to \Delta(\mathcal{S})$ is the transition function, $r(s, a) : \mathcal{S} \times 
\mathcal{A} \rightarrow \mathbb{R}$ is the reward function, $\gamma \in [0, 1)$ is the discount 
factor.
The goal is to learn a 
policy $\pi_\theta : \mathcal{S} \rightarrow \Delta(\mathcal{A})$ that maximizes expected discounted return: $J(\pi) = \mathbb{E}_{a_t \sim\pi(\cdot \mid s_t), s_{t+1} \sim P(\cdot \mid s_t, a_t)}\left[\sum_{t=0}^{\infty} \gamma^t r(s_t, a_t)\right]$. 
The $Q$-function approximates the expected discounted return from state $s$ after taking action $a$ and following the policy $\pi$: $Q^\pi(s, a) = \mathbb{E}_\pi\left[\sum_{t=0}^{\infty} \gamma^t r(s_t, a_t) \mid s_0 = s, a_0 = a\right]$. $Q$-functions are typically trained by minimizing the temporal difference (TD) loss:
$\mathbb{E}_{s, a, s' \in \mathcal{D}}(Q(s,a) - r(s,a) - \gamma \mathbb{E}_{a' \sim \pi}[Q_{\bar \phi}(s', a')])^2$.  
Notably, this loss relies on sampling from $\pi$. Since we are interested in finding $\pi$ only at test time, we instead rely on Implicit Q-learning (IQL)~\citep{kostrikov2021offline}, which allows us to learn a Q-function for near optimal policies using actions from the dataset alone. In particular, IQL minimizes loss $\mathcal{L}_{Q}(\phi)=\mathbb{E}_{(s,a,r,s')\sim\mathcal{D}}\left[(r(s,a) + \gamma V_\psi(s') - Q(s,a))^2\right]$, where $V_\psi$ is a state value function trained to regress to the upper expectile of $Q$: $\mathcal{L}_{V}(\psi)
=
\mathbb{E}_{(s,a)\sim\mathcal{D}}
[
L_2^\tau
\left(
Q(s,a)-V_\psi(s)
\right)
].$ Here $L_2^\tau$ is the expectile regression loss $L_2^\tau(u) = \left| \tau - \mathbf{1}(u < 0) \right| u^2$ and
$\tau \in (0.5, 1]$.

\paragraph{Behavior-regularized RL.}
In the \textit{offline 
RL} setting, the agent does not have access to the environment and must instead learn from a fixed 
dataset $\mathcal{D} = \{(s_i, a_t, s_i', r_i)\}_{i=1}^{|\mathcal{D}|}$ of transitions collected by 
a behavior policy $\hat{\pi}$. 
Policy optimization is then often performed by both maximizing $Q$-values while regularizing the policy towards the data distribution. Regularization is necessary since, without access to the environment, the policy can exploit out-of-distribution $Q$-values that are never corrected by real experience. To address this, methods regularize the policy from deviating too much from the behavior policy's distribution. Specifically, we consider the KL-regularized RL objective with respect to the behavior policy $\hat\pi$ where $\beta > 0$ controls the strength of the behavioral regularization:
\begin{equation}
    \label{eqn:kl-regularized-objective}
    J(\pi_\theta) = \mathbb{E}_{\tau \sim \pi_\theta}\left[\sum_{t=0}^{\infty} \gamma^t r(s_t, a_t)\right] - 
    \beta\, \mathbb{E}_{s \sim d^{\pi_\theta}(s)}\big[D_{\mathrm{KL}}\!(\pi_\theta(\cdot \mid s) \,\|\, 
    \hat{\pi}(\cdot \mid s))\big].
\end{equation}

\paragraph{Flow matching.}
Flow matching~\citep{lipman2022flow} generative models are parameterized by a time-dependent velocity field $v_\theta(x, t) : \mathbb{R}^d \times [0,1] \rightarrow 
\mathbb{R}^d$ that transports samples from a simple noise distribution $p_0 = \mathcal{N}(0, I_d)$ 
to a target data distribution $p_1 = p(x)$ via an ODE:
$
    \mathrm{d}\hat{x}_t = v_\theta(\hat{x}_t, t)\,\mathrm{d}t, \quad \hat{x}_0 \sim p_0.
$
The velocity field is trained to match the direction of linear interpolants $x_t = (1-t)x_0 + tx_1$ 
between noise and data samples via flow matching:
\begin{equation}
    \label{eq:flow matching-objective}
    \mathcal{L}_{\mathrm{FM}}(\theta) = \mathbb{E}_{t \sim \mathcal{U}[0,1],\, x_0 \sim p_0,\, 
    x_1 \sim p_1}\left[\left\| v_\theta(x_t, t) - (x_1 - x_0) \right\|^2_2\right].
\end{equation}
To use flow matching for policy training, we condition the velocity field on the state $s$ and train on dataset actions as targets. At inference time, actions are generated by integrating the ODE from  $\hat{x}_0 \sim p_0$ to $\hat{x}_1$, which serves as a valid sample from the policy distribution $\pi_\theta(a \mid s)$. Throughout the paper, we will refer to samples from the policy as ``denoised actions'', and the intermediates produced by integration as ``noisy actions''.

\section{Test-Time Gradient Guidance in RL}
\label{sec:rl-as-guidance}

Recall that our goal is to optimize the KL-regularized reward maximization objective in \cref{eqn:kl-regularized-objective}. It is well-known in the literature \citep{dayan1992feudal,abdolmaleki2018maximum,peng2019advantage} that the solution to this optimization problem is given via the closed form: 
\begin{equation}
    \label{eqn:closed-form-solution}
    \pi(a\mid s) \propto \hat{\pi}(a\mid s) \cdot \exp (Q(s, a))^{1/\beta},
\end{equation}
where we take the reference policy $\hat{\pi}(a\mid s)$ to be a flow policy that is trained via behavioral cloning loss in \cref{eq:flow matching-objective}.

We would like to transform this equation into one that tells us how to sample from high likelihood regions of $\pi$ by making use of the gradient $\nabla_aQ(s,a)$.
Unfortunately, the likelihood of our reference policy $\hat{\pi}(a | s)$ (represented by a flow matching or diffusion model) cannot be evaluated exactly. However, diffusion (and under some constraints flow matching) can be interpreted as optimizing a lower bound on the log-likelihood~\citep{KingaDiffELBO}; and diffusion models learn a score function directly. Using this perspective (assuming 
the flow corresponds to the score function $\nabla_a \log \hat{\pi}(a | s)$) we can  write the score for $\pi$ from \cref{eqn:closed-form-solution} as
\begin{equation}
    \label{eqn:score-fn-clean}
    \nabla_a \log \pi(a \mid s) = \nabla_a \log \hat{\pi}(a \mid s) + 1/\beta \cdot \nabla_a Q(s, a).
\end{equation}

Additionally, 
following prior work~\citep{fang2024diffusion}, we can extend this definition to an extended action space with noisy perturbations of actions stemming from the intermediate steps of integrating the diffusion denoising process:
\begin{equation}
    \nabla_{a_t} \log \pi(a_t \mid s) \approx \nabla_{a_t} \log \hat{\pi}(a_t \mid s) + 1/\beta \cdot \nabla_{a_t} Q(s, a_t), %
    \label{eq:guidance_noise}
\end{equation}
where $a_t$ corresponds to the partially denoised action at denoising step $t$.
This suggests that in order to sample from the improved policy $\pi$, we can modify the denoising process to integrate over the score function of the reference policy plus a guidance term $\nabla_{a_t} Q(s, a_t)$. %
A higher weight on the guidance term leads to a policy that is less constrained by the reference policy, and maximizes the $Q$ function to a higher degree. This is analogous to performing classifier guidance \citep{dhariwal2021diffusion} with a learned Q function replacing the classifier.

\paragraph{Problems with na\"ive Q gradient guidance.}
The guidance term in \cref{eq:guidance_noise} is the gradient of the $Q$-function at noisy actions $a_t$. This gradient can be unreliable since the critic $Q(s, a)$ is only trained on the denoised action space and querying it at out-of-distribution noisy actions may require the gradient of the Q-function to be correct far from its training data, which is not generally guaranteed. We therefore refer to $\nabla_{a_t} Q(s, a_t)$ as the ``OOD gradient''.
Taking the gradient at noisy action $a_t$ is also problematic because $a_t \in \mathbb{R^{|A|}}$ might not even correspond to a valid action in $\mathcal{A}$.

\begin{figure}[t]
    \centering
    \includegraphics[width=\linewidth]{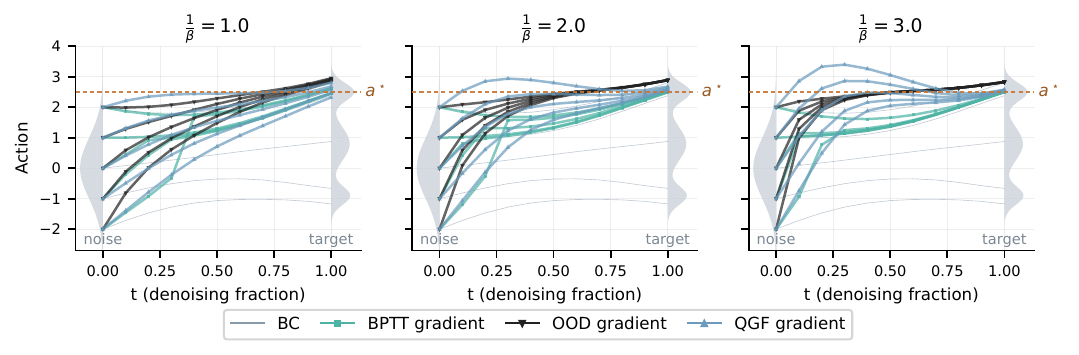}
    \vspace{-0.5cm}
    \caption{Illustrative example of 1D denoising process mapping Gaussian noise to a tri-modal distribution, with $Q$ defined as negative L2 distance to the optimal action $a^*$. We compare the base BC flow and three critic-gradient guidance methods (BPTT, OOD, QGF) across three guidance weights. While BPTT and \methodname{} converge to $a^*$, \textbf{guidance with the OOD gradient $\nabla_{a_t} Q(s, a_t)$ does not result in the optimal solution}. Further, \cref{fig:analytical_guidance_bptt} shows \textbf{BPTT can be highly unstable}.}
    \label{fig:analysis-1d-denoising}
\end{figure}

\begin{wrapfigure}{rh}{0.42\linewidth}
\vspace{-0.5cm}
\centering
    \includegraphics[width=\linewidth]{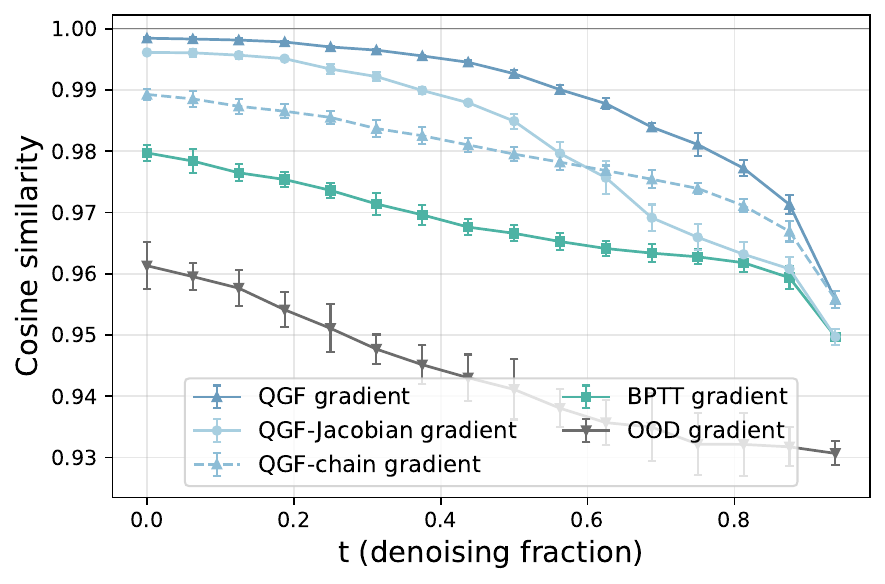}
    \vspace{-0.5cm}
    \caption{Sensitivity of different gradient estimators to noise in the action space: for each gradient estimator $G$, the plot shows the cosine similarity between $G(s, a_t)$ and $G(s, a_t + \epsilon)$. Our proposed gradient estimator has the least variance and least sensitivity to noise. Averaged over $20$ tasks and $4$ seeds.}
    \label{fig:gradient-estimator-variance}
    \vspace{-1em}
\end{wrapfigure}

Alternatively, we could consider ways to query Q on denoised actions only. Since flow matching deterministically maps each noisy action $a_t$ to a denoised action $a_1$ by integrating through the ODE (\ie $a_1 = \mathrm{ODE }(a_t) = a_t + \int_t^1 v_\theta(a_\tau,\tau) d\tau$), we can interpret the Q value of a noisy action to be the Q value of its denoised version:
\begin{equation}
    \label{eqn:def-noisy-q}
    Q(s, a_t) := Q(s, \text{ODE }(a_t)).
\end{equation}
Thus, a more principled way would be to use $\nabla_{a_t} Q(s, \text{ODE }(a_t))$. Computing this gradient requires back propagation through the denoising process, which we call the ``BPTT gradient''.
While more principled, computing this gradient is expensive. 
Furthermore, we find that the BPTT gradient is very sensitive to noise and has a high variance. This effect is clearly visible in \cref{fig:gradient-estimator-variance}, where we plot the cosine similarity between the gradient evaluated at $a_t$ and a slightly perturbed noisy action $a_t + \epsilon$, \ie $\cos(G(s, a_t), G(s, a_t + \epsilon))$ where $G$ corresponds to the estimated gradient. Values closer to $1$ correspond to gradient estimators that are less sensitive to noise.

\paragraph{Illustrative example of suboptimal $Q$-gradient guidance.}
We show a simple illustrative example in which using either of the above described gradient estimators results in suboptimal denoised actions. In \cref{fig:analysis-1d-denoising}, we visualize a 1D denoising process where a flow matching model tries to map input noise distribution to a tri-modal action distribution. The ``Q-function'' in this didactic example is the negative squared distance to the optimal action. We visualize the analytical solution to the base behavioral cloning (BC) flow and the solution when it is guided by three different gradients: the OOD gradient $\nabla_{a_t} Q(s, a_t)$, the BPTT gradient $\nabla_{a_t} Q(s, \mathrm{ODE}(a_t))$, and the approximate gradient we propose in \cref{sec:method}. As we can see, regardless of the guidance weight, using the OOD action gradient always misguides the flow and ``bias'' it to a suboptimal action after denoising. \cref{sec:results} also corroborates this illustrative example and shows that test-time guidance with the OOD gradient (following ~\citep{jang2025q}) results in poor performance. While the BPTT gradient is less ``biased'', it is still highly sensitive to noise (see \cref{fig:gradient-estimator-variance}) and therefore leads to instability for policy optimization, as we find in \cref{fig:analytical_guidance_bptt}. On the other hand, the gradient estimator that we propose below, \methodname{}, has much smaller variance (see \cref{fig:gradient-estimator-variance}) and results in stable optimization towards the optimal action.

\section{\methodexplain}
\label{sec:method}

\begin{wrapfigure}{rh}{0.45\linewidth}
\centering
    \includegraphics[width=\linewidth]{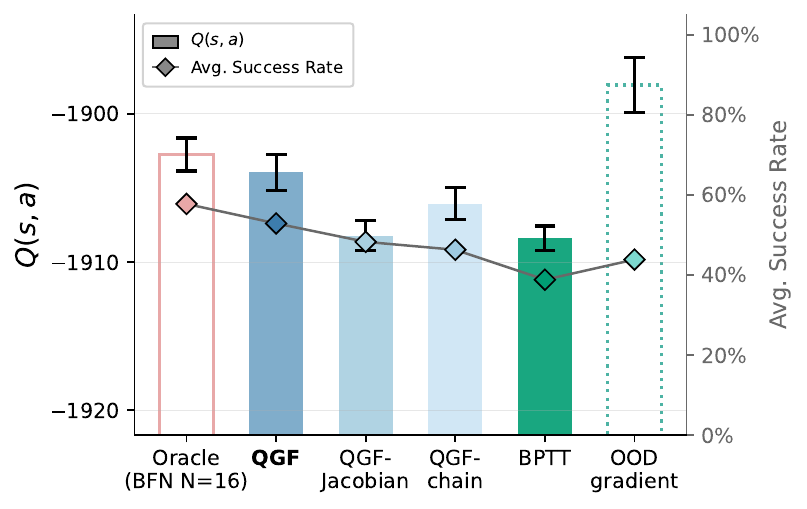}
    \vspace{-1.7em}
    \caption{$Q$-values of the denoised action with different gradient guidance, which roughly correlate with performance. Averaged over $20$ OGBench environments. \textbf{\methodname{} is the best gradient-based optimizer, getting close to the performance of the best-of-n oracle.}}
    \label{fig:better-optimizer}
    \vspace{-1em}
\end{wrapfigure}

In this section, we will show that we can derive a gradient estimator that: i) does not query the critic on OOD inputs, ii) is cheaper than BPTT through ODE integration, and iii) has lower variance and consequently more effectively optimizes actions to maximize $Q$-values.
We start by noting that, instead of fully integrating the denoising ODE, we can obtain a cheap, first-order approximation to the ODE solution by taking a single large Euler integration step~\citep{bansal2024universal, efron2011tweedie}, following the local velocity field at action $a_t$ all the way to a denoised action:
\begin{equation}
    \label{eqn:one-step-denoising}
    \hat{a}_1 = a_t + v_\theta(s, a_t, t) \cdot (1-t),
\end{equation}
where $v_\theta(s, a_t, t)$ is the flow velocity function of the reference policy $\hat{\pi}(a\mid s)$ at time step $t$. 
As we will show in \cref{fig:better-optimizer}, using this first-order approximation is not just a convenience. Perhaps surprisingly, it is actually \emph{more} effective than running the full denoising process because it is less constrained to the exact dataset distribution.
From this, we can approximate the ground truth gradient as 
\begin{equation}
    \label{eqn:jacobian}
    \nabla_{a_t} Q(s, a_1) \approx \nabla_{a_t} Q(s, \hat{a}_1) = \left(\frac{\partial \hat{a}_1}{\partial a_t}\right)^\top \nabla_{\hat{a}_1} Q(s, \hat{a}_1),
\end{equation}
which is a product between the gradient of $Q$ and the Jacobian of the denoised action with respect to the noisy action, $J=\frac{\partial \hat{a}_1}{\partial a_t}$.
Empirically, we find that $J$ can be ill-behaved since it requires differentiating through $v_\theta(s, a_t, t)$, and replacing it with the identity entirely ($J \approx \hat{J} :=I$) yields better performance, effectively computing
\begin{equation}
    \label{eqn:gradient-approx}
    \nabla_{a_t} Q(s, a_1) \approx \hat{J}^\top \: \nabla_{\hat{a}_1} Q(s, \hat{a}_1) \quad \text{where} \quad {\hat{a}_1 = a_t + v_\theta(s, a_t,t) \cdot (1-t), \: \hat{J}=I}.
\end{equation}

\newcommand{\algcomment}[2][gray]{%
{\color{#1}\hfill$\blacktriangleleft$~\footnotesize\textit{#2}}}

\begin{wrapfigure}{r}{0.5\linewidth}
\vspace{-\baselineskip}
\begin{minipage}{\linewidth}
\captionof{algorithm}{\methodname{} (Q-Guided Flow) inference.}
\label{alg:qgf}
\vspace{-0.3\baselineskip}
\begin{tabular}{p{\linewidth}}
\toprule
\textbf{Inference time} \\
\midrule
\textbf{Input:} state $s$, ref. flow $v_\theta$, $Q$, guid. weight $\frac{1}{\beta}$, $\delta=1/T$ \\[4pt]
$a_0 \sim \mathcal{N}(0, I)$ \\
\textbf{for} $t = 0, \delta, 2 \delta \dots, 1-\delta$ \textbf{do} \algcomment{flow denoising} \\
\quad $\hat{a}_1 \gets a_t + (1-t)\,v_\theta(s, a_t, t)$ \\
\quad $g \gets \nabla_{\hat{a}_1} Q(s, \hat{a}_1)$ \algcomment[ourblue]{\methodname{} grad.\ est.} \\
\quad $a_{t+\delta} \gets a_t + \delta \cdot \left(v_\theta(s, a_t, t) + \frac{1}{\beta}\,g\right)$ \algcomment{Q guidance} \\
\textbf{end for} \\
\textbf{return} $a_1$ \\
\bottomrule
\end{tabular}
\end{minipage}
\end{wrapfigure}
In summary, we use a first-order approximation to denoise action $a_t$, and take the critic gradient at the denoised action. \methodname{} uses this $Q$ gradient to guide the denoising of a reference policy during inference time, essentially adding a weighted value of this gradient to the velocity of the reference policy to steer it towards more optimal actions. The full algorithm is described in \cref{alg:qgf}. Two design choices in \methodname{} may appear to be crude approximations: we drop the Jacobian entirely and use a first-order approximation of the denoised action. Both look like approximations one would only tolerate for efficiency. Surprisingly, we find that neither is merely a compromise and both outperform their more ``exact'' counterparts. In the analysis below, we show that these choices yield a lower-variance gradient estimator that is better at optimizing $Q$-values.

\paragraph{Analysis of the Jacobian.}
We shall refer to \cref{eqn:gradient-approx} as the \methodname{} estimator and \cref{eqn:jacobian} as the \methodname{}-Jacobian estimator. 
Calculating the Jacobian essentially requires taking partial derivative through $v_\theta$, and can be ill-behaved at times, e.g., in early steps when \cref{eqn:one-step-denoising} is a crude approximation of the ground truth denoising process.
We empirically find two ways to address this: regularizing the Jacobian (see \cref{fig:qgf_reguarlized_jacobian}) or replacing it with the identity. 
Indeed, \cref{fig:gradient-estimator-variance} shows that including the Jacobian in the gradient estimator can make it more sensitive to noise in the input.
This high sensitivity to noise in turn makes \methodname{}-Jacobian a worse ``optimizer'' for $Q$-values than \methodname{}: we can view each gradient estimator that guides flow denoising as an ``optimizer'' that tries to optimize for high $Q$ value actions, which we find to roughly correlate with performance.
\cref{fig:better-optimizer} shows the $Q$ value of the final denoised actions when following different gradient guidance during denoising, and shows \methodname{} achieves a higher $Q$ value than \methodname{}-Jacobian.
We compare both $J=\frac{\partial \hat{a}_1}{\partial a_t}$ and $J \approx I$ in our experiments below, and find that in fact not using the Jacobian performs better (see \cref{fig:main-results}), especially on hard environments (see \cref{fig:dqc-results}). We note that there is a large body of work that shows that such approximate gradients are often sufficient~\citep{lillicrap2014random, nokland2016direct, jaderberg2017decoupled, launay2020direct, mcallister2026finite}.

\paragraph{First-order approximation.}
To analyze the impact of the first-order approximation, we compare to a variant where the denoised action is obtained by following the entire denoising chain: $a_1 = \text{ODE}(a_t)$. We compare both gradient estimators ($\nabla_a Q(s, a)|_{a=\hat{a}_1}$ v.s. $\nabla_a Q(s, a)|_{a=a_1}$) without the Jacobian (\ie $J \approx I$) since that leads to better performance and avoids the instability of BPTT. We will refer to this variant as the \methodname{}-chain estimator. 
Similarly, we find that \methodname{}-chain is more sensitive to noise in \cref{fig:gradient-estimator-variance}, making it also a worse optimizer for $Q$-values in \cref{fig:better-optimizer}.
Indeed, we find in \cref{fig:qgf_reguarlized_jacobian} that \methodname{}-chain underperforms \methodname{}.
We hypothesize this is due to \methodname{} being better at mode selection: following the full denoising process of the base BC flow restricts the denoised action to cover the full dataset distribution, while \methodname{} allows deviation from the exact dataset distribution and allows the flow to choose only certain modes of the dataset distribution. This ability to choose good modes enables \methodname{} to guide a base BC policy to select better actions at test time.

In addition, \cref{fig:better-optimizer} also shows \methodname{} to be a better Q-optimizer than all the other gradient-based estimators except for the OOD gradient, which we find to exploit the $Q$ function with OOD actions and does not lead to good performance (see \cref{apdx:better-optimizer} for a detailed discussion). Overall, our analyses shed some light on why \methodname{} is a particularly effective gradient-estimator for test-time guidance.

\paragraph{Implementation details.}
\methodname{} is a test-time RL algorithm, using an approximated critic gradient in \cref{eqn:gradient-approx} to guide action denoising. The algorithm is agnostic to the training of the reference policy and the critic. For simplicity, in this paper, we train the reference policy with behavioral cloning and the value function with IQL~\citep{kostrikov2021offline}. We purposely chose an in-sample value-learning algorithm for critic training so that we can fully decouple value learning and policy extraction. See \cref{sec:different-critics} for using \methodname{} with other types of value functions.

\section{Experimental Results}
\label{sec:results}

We conduct experiments to study how well \methodname{} is able to optimize the policy at test-time on a range of long-horizon challenging manipulation tasks. We focus on manipulation domains because offline datasets in these environments typically have lower state-action coverage and are therefore harder for RL algorithms.
Concretely, we study the following research questions: \textbf{(1)} Can \methodname{} optimize the policy well compared to other RL methods? \textbf{(2)} Can more test-time compute increase the performance for \methodname{}? \textbf{(3)} Can \methodname{} be scaled to solve hard tasks in goal-conditioned RL and utilize larger model architectures? \textbf{(4)} Can \methodname{} work with different types of critics?

\subsection{Experimental Setup}
We consider seven different environments (5 tasks each) from OGBench~\citep{park2024ogbench} in the offline RL setting: \texttt{scene}, 
\texttt{puzzle-4x4} (\texttt{p44}), 
\texttt{puzzle-4x5} (\texttt{p45}), 
\texttt{puzzle-4x6} (\texttt{p46}), 
\texttt{cube-triple} (\texttt{c3}), 
\texttt{cube-quadruple} (\texttt{c4}), 
\texttt{cube-octuple} (\texttt{c8}). We follow the action chunking setting~\citep{li2025reinforcement} with a chunk size of $h=5$.
In this setting, the policy outputs a sequence of $h$ actions and executes them one by one in the environment. This high-dimensional action space induces a much more complex distribution compared to using single-step atomic actions, and therefore makes an ideal testbed to compare policy extraction methods. Since learning from the default OGBench datasets (ranging from $1$M to $3$M transitions in size) has saturated performance for offline RL with action chunking~\citep{li2026q}, we use the datasets with $100$M transitions~\citep{li2025reinforcement}. These larger datasets allow for a bigger jump in performance (compared to the data collection policy). 
To stress test our method we also consider the $1$B transitions dataset for \texttt{p46} and \texttt{c8}. Finally, we consider a challenging sparse reward setting for the harder puzzles environments (\texttt{p45} and \texttt{p46}).

\subsection{Baselines}
We compare \methodname{}, as well as its variant \methodname{}-Jacobian (which includes the Jacobian from \cref{eqn:jacobian}), to a representative set of baselines from prior work. We group the baselines into two categories: 

\textbf{(1) Test-time methods}, which use a policy pre-trained with BC and a learned critic to optimize sampled actions at test time: \textbf{BFN}~\citep{li2025reinforcement, hansen2023idql} samples $N$ actions from the actor and picks the highest value one according to the critic; \textbf{GradStep} improves fully denoised actions at test time by taking a few steps in the gradient direction~\citep{mark2024policy, xureinforcement}; \textbf{QFQL}~\citep{jang2025q} guides each denoising step with the OOD gradient in \cref{sec:rl-as-guidance} (\ie $\nabla_{a_t} Q(s, a_t)$); 
\textbf{BPTT}, a novel baseline inspired by DQL~\citep{wang2022diffusion}, uses the BPTT gradient estimator for guidance;
\textbf{CFGRL}~\citep{frans2025diffusion}, which trains a policy by conditioning it on advantage values and samples with classifier free guidance at test time;
\textbf{RobustQ}, a novel baseline inspired by robust classifiers~\citep{santurkar2019image, kawar2022enhancing, tsipras2018robustness, ilyas2019adversarial} in classifier guidance, takes gradients on a ``robust'' $Q$ function $Q(s, a_t, t)$ trained with noisy actions and denoising time steps as inputs.

\textbf{(2) Training-time methods}, which train the policy to maximize $Q(s, a)$ during training: \textbf{FQL}~\citep{park2025flow} learns a distilled one-step policy for ascending on the critic gradient; \textbf{EDP}~\citep{kang2023efficient} uses a first-order Euler approximation of the denoised action; \textbf{QAM}~\citep{li2026q} uses adjoint matching to replace back propagating through the denoising process; \textbf{DAC}~\citep{fang2024diffusion} trains a diffusion model to match base BC distribution tilted by $\exp(Q(s, a))$; \textbf{QSM+BC} add BC regularization to QSM~\citep{psenka2023learning}, which uses a diffusion model to match the gradient of $Q(s, a)$.

See a more detailed description of these methods in \cref{apdx:baselines-description}. To ensure a fair comparison of policy extraction methods, the same critic $Q(s, a)$ is used across all methods. We use the IQL~\citep{kostrikov2021offline} Q-value function described in the previous section.
See \cref{sec:different-critics} for details on different critic formulations.

\begin{figure}[t]
    \centering
    \includegraphics[width=\linewidth]{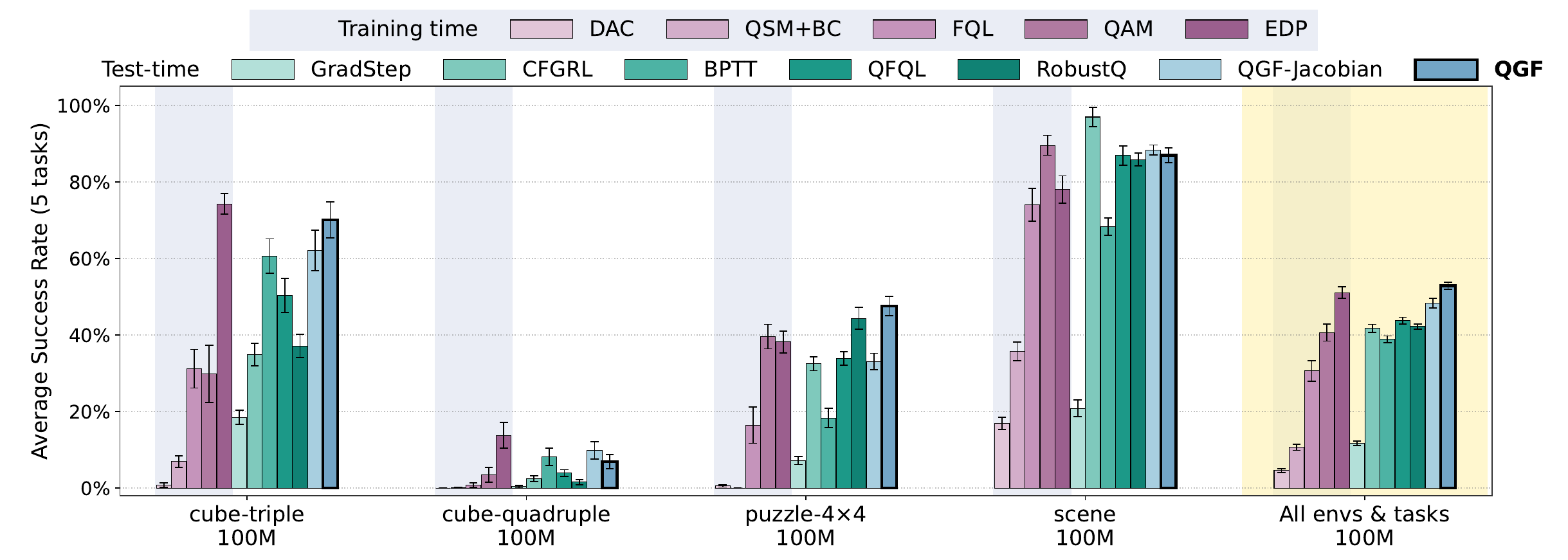}
    \vspace{-0.5cm}
    \caption{Offline RL performance at $500$k training steps (20 tasks, 10 seeds): \methodname{} beats all previous test-time methods and is competitive with the best training-time method. See task breakdown in \cref{fig:full-result-no-bfn}.}
    \vspace{-0.3cm}
    \label{fig:main-results}
\end{figure}

\subsection{Comparing \methodname{} with Test-Time and Train-Time Baselines}
\label{sec:result-main}

\begin{figure}[t]
    \begin{minipage}[t]{0.6\linewidth}
        \centering
        \includegraphics[width=\linewidth]{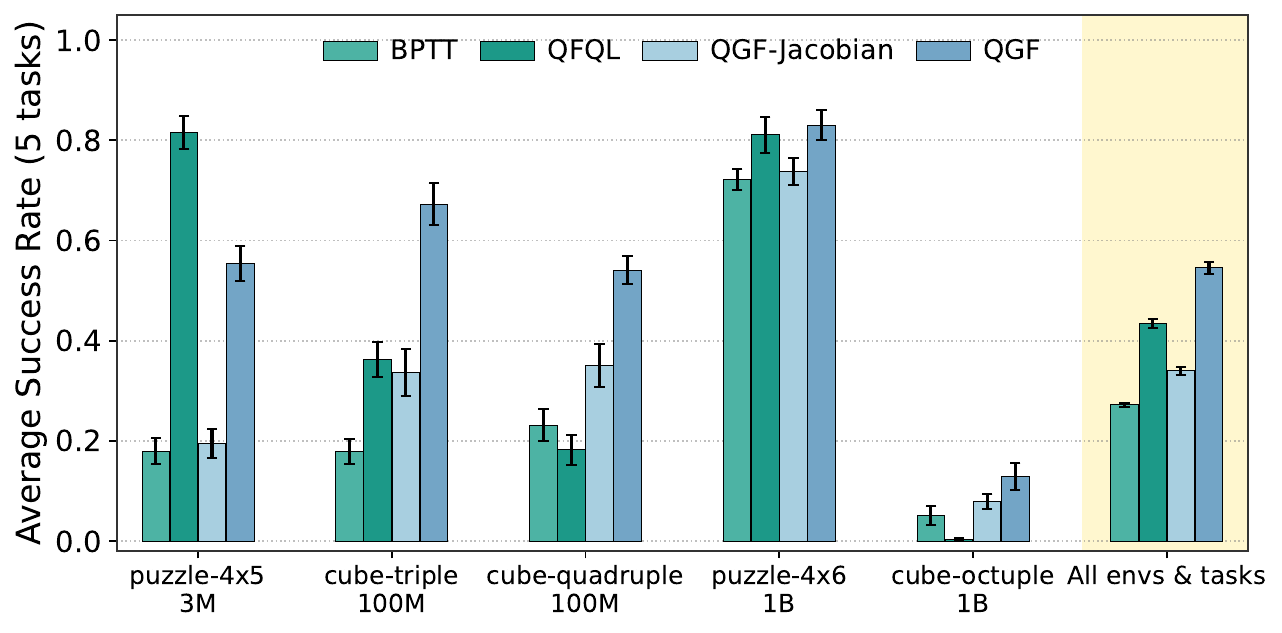}
        \vspace{-0.5cm}
        \caption{\textbf{Offline goal-conditioned RL performance} at $1$M training steps (25 tasks, 10 seeds): While \methodname{} underperforms the best baseline on the simplest task, it is consistently the best performing method for the harder tasks, showing that the gradient estimator scales well to long horizon tasks. See task breakdown in \cref{fig:full-result-dqc}.}
        \label{fig:dqc-results}
    \end{minipage}
    \hfill
    \begin{minipage}[t]{0.35\linewidth}
        \centering
        \includegraphics[width=\linewidth]{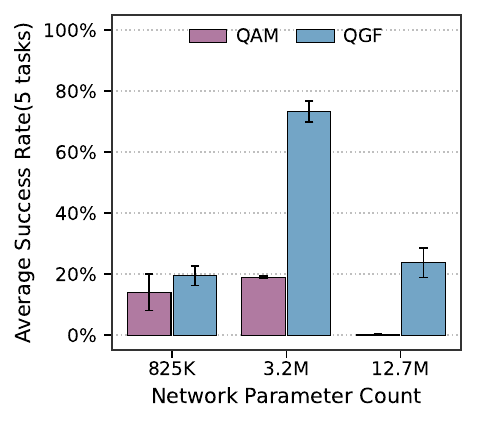}
        \vspace{-0.5cm}
        \caption{\textbf{Scaling model size of \methodname{}}: compared to training-time methods such as QAM~\citep{li2026q}, \methodname{} benefits much more from increasing the model size. Averaged over 5 tasks in \texttt{c3}.}
        \label{fig:result-model-size-ablation}
    \end{minipage}
\end{figure}

\cref{fig:main-results} shows that \methodname{} outperforms previous gradient-based guidance algorithms (QFQL, BPTT, RobustQ) significantly. 
This shows the strength of our proposed gradient estimator in \cref{eqn:gradient-approx}. We speculate that this is due to its lower variance as shown in \cref{fig:gradient-estimator-variance}. \methodname{} also significantly outperforms CFGRL and GradStep, indicating that gradient guidance during each denoising step is more effective than classifier free guidance, and than using the critic gradient to change the fully denoised action only.
In addition, \methodname{} also outperforms \methodname{}-Jacobian, showing that not using the Jacobian in the gradient estimator improves performance, which is also confirmed later in \cref{fig:dqc-results} on harder OGBench tasks.

Compared to methods that explicitly train the policy to maximize Q functions (which we call training-time methods), \methodname{} only trains the policy with a standard flow matching loss, and uses the Q function gradient to guide action sampling at test-time. Training-time methods typically require tuning a hyperparameter that balances reward maximization against a behavioral constraint (such as the weight for the behavioral constraint in EDP), and good performance usually requires extensive tuning of this hyperparameter. In comparison, \methodname{} does not need to tune such a parameter during training. Instead, its behavior can be adjusted at test time without needing to re-train a policy. However, we still compare to representative training-time methods (FQL, EDP, QAM, DAC, QSM+BC). Surprisingly, \methodname{} outperforms most training-time methods and is competitive and slightly better than the best. This demonstrates that test-time RL is a powerful policy optimization technique. We note that the best training-time baseline method, EDP, also uses a first-order approximation of the denoised action, showing the effectiveness of this approximation.

\begin{figure}[t]
    \vspace{-0.5em}
    \centering
    \begin{minipage}{0.45\linewidth}
        \centering
        \includegraphics[width=\linewidth]{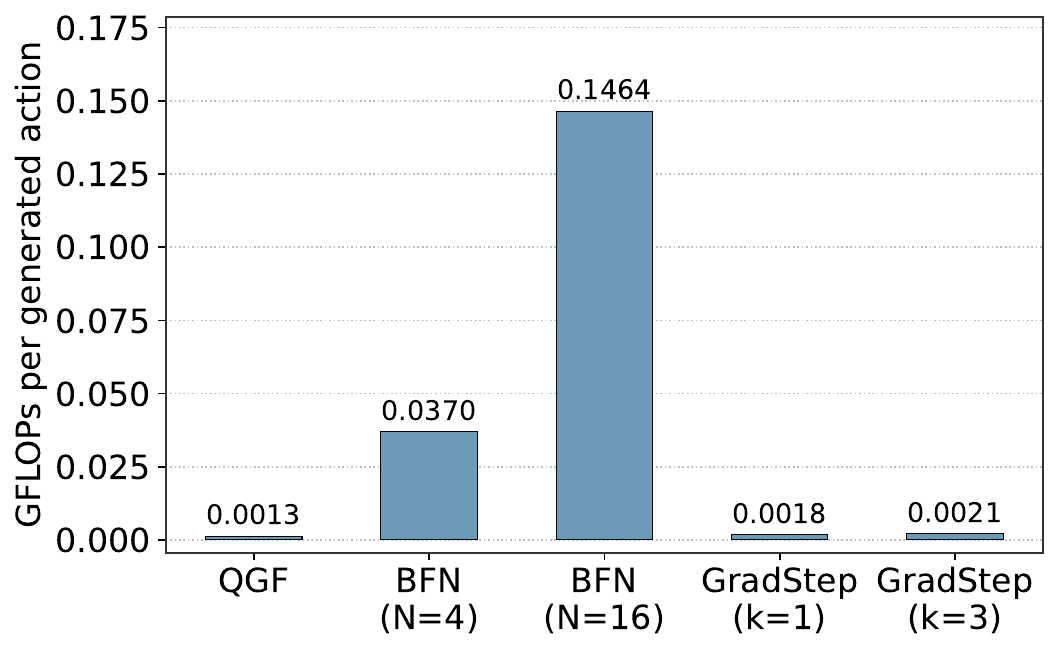}
        
        \caption{Compute requirements for different test-time methods: BFN needs orders of magnitude more FLOPs compared to \methodname{} and GradStep.}
        \label{fig:flops-comparison}
    \end{minipage}
    \hfill
    \begin{minipage}{0.52\linewidth}
        \centering
        \includegraphics[width=\linewidth]{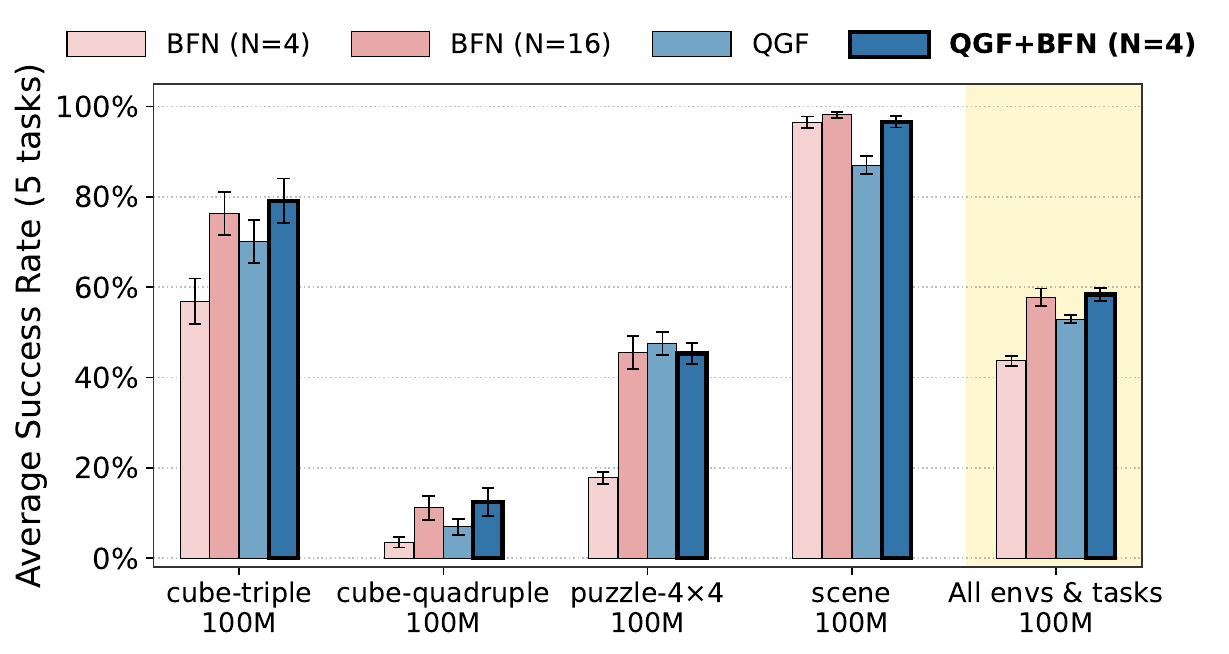}
        \vspace{-0.5cm}
        \caption{Offline RL performance at 500k training steps (20 tasks, 10 seeds) evaluated with best-of-N sampling. \methodname{} outperforms BFN (N=4), and \methodname{}+BFN matches BFN with much higher test-time compute budget. See task breakdown in \cref{fig:full-result-bfn}.}
        \label{fig:result-bfn}
    \end{minipage}
    \vspace{-0.4cm}

\end{figure}

\paragraph{Scaling test-time compute.}
One effective way to improve policy performance when additional test-time compute is available is to incorporate best-of-N (BFN) sampling. In BFN, multiple samples are generated from a reference policy and we select the sample with the highest Q-value. This has proven to be an effective policy extraction method for both robotic control~\citep{li2025reinforcement, hansen2023idql, nakamoto2024steering} and language modeling~\citep{snell2024scaling}.
However, BFN is very expensive since it requires rolling out the entire denoising process multiple times to select a promising action. In fact, \cref{fig:flops-comparison} shows that BFN is orders of magnitude more expensive than \methodname{} and other test-time methods that we consider.
In \cref{fig:result-bfn}, we compare \methodname{} to methods that utilize best-of-N sampling. We introduce a variant of our method, \methodname{}+BFN, which samples $N$ actions from \methodname{} and picks the highest value one ranked by the critic. As we can see, while BFN (N=4) alone uses drastically more test-time compute than \methodname{}, it actually performs worse, indicating that \methodname{} is a more effective value maximizer on its own. With even more test-time compute, BFN (N=16) catches up, achieving similar and slightly better performance than \methodname{}. However, when given more compute, \methodname{}+BFN is able to beat \methodname{}, and match BFN (N=16) with a smaller test-time compute budget, only requiring $4$ samples instead of $16$.

\subsection{Scaling \methodname{} to Harder Tasks and Larger Models}

To test whether our method can solve even harder tasks than those in \cref{sec:result-main}, we also evaluate under the goal-conditioned RL setting instead of single-task RL.
We evaluate \methodname{} and its variant \methodname{}-Jacobian with the two most relevant baselines that use critic gradients at test-time (BPTT, QFQL) on the five most challenging environments in OGBench.
Since these are hard long horizon tasks, we use the corresponding larger datasets that support learning~\citep{park2025horizon}.
In this experiment, we train IQL value functions with DQC~\citep{li2025decoupled} since the tasks have a very long horizon. \cref{fig:dqc-results} shows that on the easiest task, \texttt{p45}, \methodname{} actually underperforms QFQL. However, as we move to harder tasks,
\methodname{} is consistently the best-performing method, showing that our critic gradient estimator in \cref{eqn:gradient-approx} scales well on hard long horizon tasks. It is also worth noting that on the hardest tasks, \methodname{} consistently outperforms the \methodname{}-Jacobian variant. This shows that it is beneficial to use a lower-variance gradient estimator that does not include the Jacobian.

To evaluate whether \methodname{} scales well with model size, we compare \methodname{} at different model sizes to a training-time baseline, QAM~\citep{li2026q}. \cref{fig:result-model-size-ablation} shows the performance for the two agents using three model sizes for both the actor and the critic, and we find that when scaling the model size from 800k parameters to 3.2M parameters, QAM does not improve while \methodname{} experiences a nearly $4\times$ jump in performance. 
When we increase the model size past a certain threshold (e.g., 12.7M parameters), both methods experience overfitting, though \methodname{} suffers less while QAM results in a policy that is unable to complete the task.

\subsection{\methodname{} with Different Critics}
\label{sec:different-critics}

\begin{wrapfigure}{rh}{0.48\linewidth}[h]
\vspace{-0.5cm}
\centering
    \centering
    \vspace{-0.3em}
    \includegraphics[width=\linewidth]{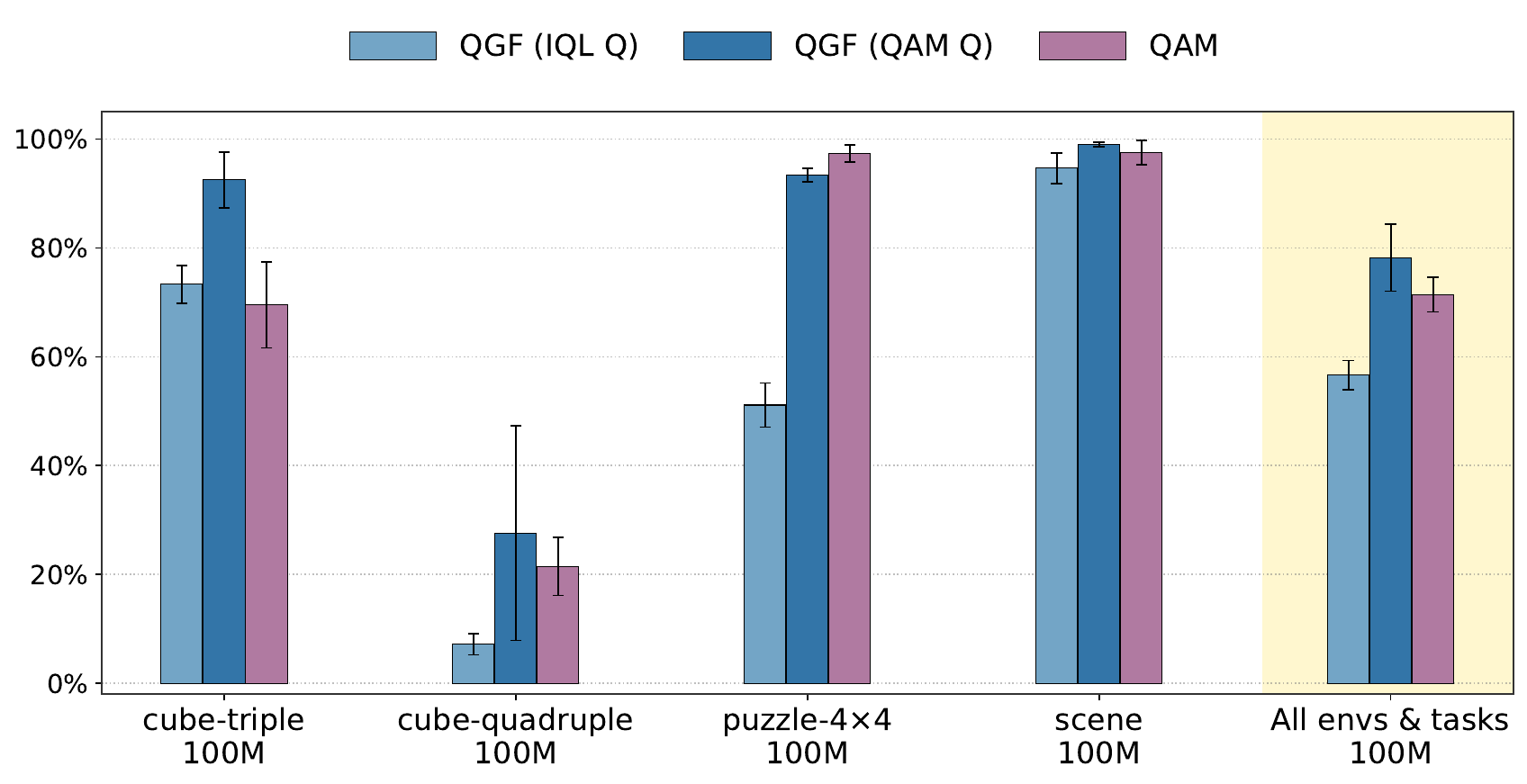}
    \caption{\methodname{} can work with different types of critics and performs better when critics are better (20 tasks, 4 seeds). See task break down in \cref{fig:qam-value-functions-full}.}
    \vspace{-1em}
    \label{fig:qam-value-functions}
\end{wrapfigure}
All the above results compare \methodname{} and all baseline methods with an in-sample learning critic using IQL~\citep{kostrikov2021offline}. 
Here, we ask whether \methodname{} can also work with critics trained with actions sampled from the policy via $Q$ bootstrapping~\cite{lillicrap2015continuous, fujimoto2018addressing, haarnoja2018soft} since these critics perform extremely well when tuned carefully~\cite{park2024value}. We choose to train the critic with action sampling from QAM~\citep{li2026q} because it is a top-performing method.
\cref{fig:qam-value-functions} shows that \methodname{} using the QAM-based $Q$ function performs much better than IQL-based \methodname{}, and also performs better than QAM with $Q$ bootstrapping.
This shows that \methodname{} is an extremely effective policy extraction method that can work with different types of $Q$ functions.

\section{Conclusion}

In this paper, we introduced \methodname{}, a test-time RL algorithm that improves flow behavioral cloning policies by guiding each denoising step with a critic gradient evaluated at an approximated clean action. Empirically, \methodname{} outperforms prior test-time RL methods and is competitive with the best training-time RL baselines. This suggest that test-time gradient guidance is a practical and scalable alternative to actor-critic policy optimization in continuous control. There are several promising directions for future work. First, we focus on the offline RL setting, and an immediate next step is to explore \methodname{} in the online RL setting. Second, we focus on policy extraction assuming a reasonably accurate critic. Developing methods to train more accurate critics is an important research question. Another interesting direction is to develop policy extraction techniques that are robust to inherently noisy critics.

\section{Acknowledgements}
This research was partly supported by AFOSR FA9550-22-1-0273 and DARPA ANSR. This research used the Savio computational cluster resource provided by the Berkeley Research Computing program at UC Berkeley. We would like to thank Seohong Park, Chung Min Kim, and Kai Nelson for helpful discussions and feedback on early drafts of the paper. 

\section{Author Contributions}
ZZ led the project, carried out the major experiments, and led the writing. AP ran substantial experiments that helped iterate and analyze the method, and also assisted with writing. CX contributed to early project ideation and discussions, and helped with experiments, method analysis, and writing. QL ran the goal-conditioned RL experiments and, together with TS, KF, and SL, provided valuable guidance and advice throughout the project and contributed to writing.

\bibliography{ref}
\bibliographystyle{plain}

\newpage
\appendix

\section{Extra Analysis on BPTT Gradient Guidance}
\begin{figure}[H]
    \centering
    \includegraphics[width=\linewidth]{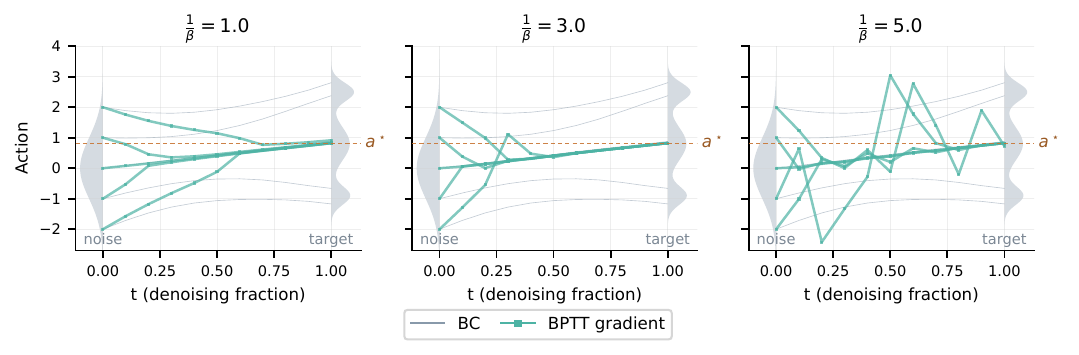}
    \vspace{-0.5cm}
    \caption{We find that the BPTT gradient can be unstable for higher guidance weights and certain target distributions. }
    \label{fig:analytical_guidance_bptt}
\end{figure}

Here we offer another example of our 1-D illustrative denoising example described in \cref{sec:rl-as-guidance}. In this example, we find that the BPTT gradient can be extremely unstable, as shown in \cref{fig:analytical_guidance_bptt}. This suggests that using such gradient guidance could lead to suboptimal results, as confirmed by our experiments in \cref{fig:main-results} and \cref{fig:dqc-results}.

\section{\methodname{} Variants}
\label{apdx:qgf-variants}

\begin{figure}[H]
    \centering
    \includegraphics[width=1\linewidth]{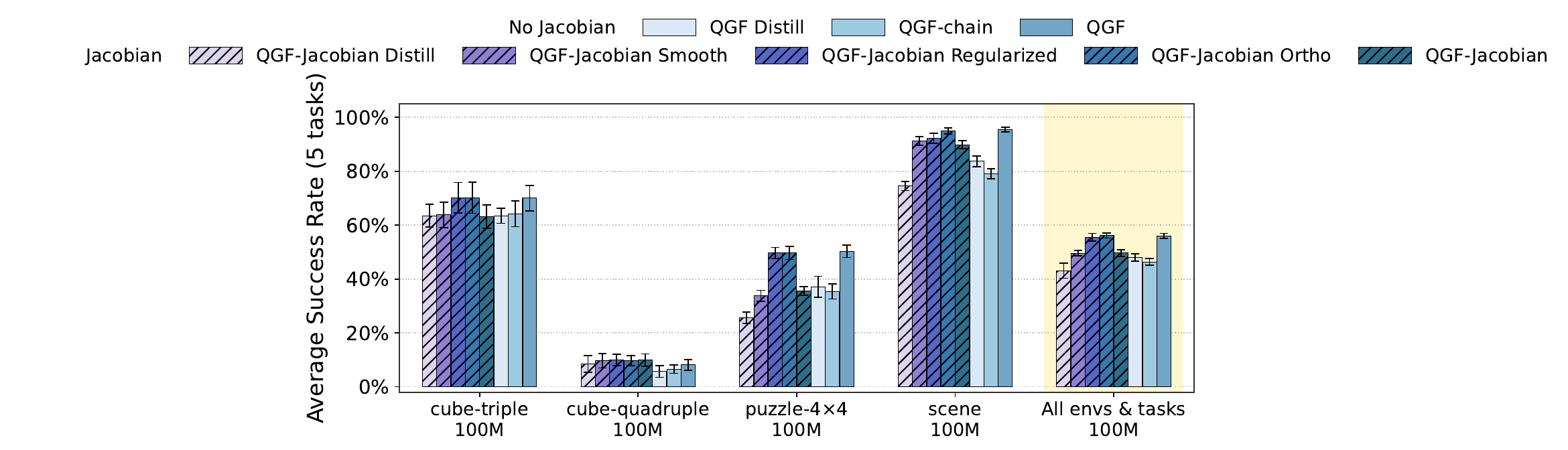}
    \caption{Offline RL performance at 500k training steps (20 tasks, 10 seeds) of different \methodname{} variants. Methods that apply the Jacobian are shaded. See task specific break down in \cref{fig:qgf_reguarlized_jacobian_full_results}.}
    \label{fig:qgf_reguarlized_jacobian}
\end{figure}

\begin{figure}[h]
    \centering
    \includegraphics[width=1\linewidth]{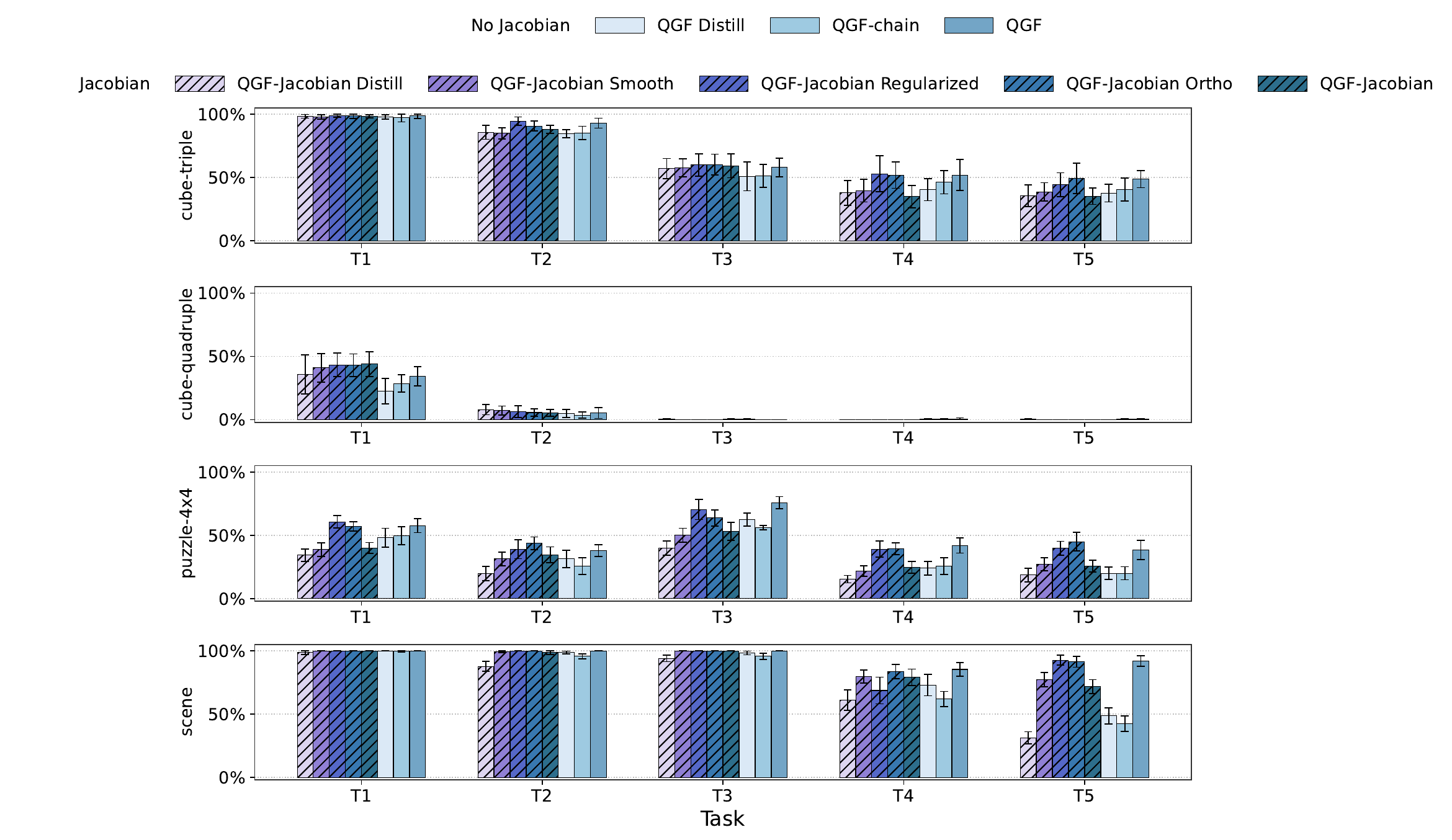}
    \caption{Task breakdown of \cref{fig:qgf_reguarlized_jacobian}: Offline RL performance at 500k training steps (20 tasks, 10 seeds).}
    \label{fig:qgf_reguarlized_jacobian_full_results}
\end{figure}

We present in \cref{fig:qgf_reguarlized_jacobian} the comparison between \methodname{} and many variants of \methodname{}. See the task specific breakdown in \cref{fig:qgf_reguarlized_jacobian_full_results}. \methodname{}, \methodname{}-Jacobian, and \methodname{}-chain are explained in \cref{sec:method}, and we detail the other variants below:

\paragraph{\methodname{} Distill / \methodname{}-Jacobian Distill.}
Instead of using the first-order approximation in \cref{eqn:one-step-denoising} to approximate the fully denoised action, one can also train a ``distilled'' velocity field to denoise noisy actions. For any noisy action $a_t$ at timestep $t$, we can train a velocity field $v_{d}(s, a_t, t)$ with loss:
$$||v_{d}(s, a_t, t) - \text{ODE}(a_t)||^2,$$
where $\text{ODE}(a_t)$ represent the full denoising chain of the reference policy ($\text{ODE }(a_t) = a_t + \int_t^1 v_\theta(a_\tau,\tau) d\tau$). Then, we approximate the clean action with $\hat{a}_1 = v_d(s, a_t, t)$, and approximate the gradient with
$$\nabla_{a_t} Q(s, \hat{a}_1) = J^T \: \nabla_{\hat{a}_1} Q(s, \hat{a}_1).$$
\methodname{}-Jacobian Distill would calculate the Jacobian $J=\frac{\partial \hat{a}_1}{\partial a_t}$ by taking derivative through $v_d$; \methodname{} Distill would simply set $J \approx \hat{J}=I$.

\paragraph{\methodname{}-Jacobian Smooth.}
Since we found in \cref{fig:gradient-estimator-variance} that \methodname{}-Jacobian has higher ``variance'' than \methodname{}, we explore whether we can reduce the variance of the \methodname{}-Jacobian estimator with some ``smoothing'' method.
Here we also use the one-euler-step approximation of $\hat{a}_1 = a_t + (1-t)v_\theta(s, a_t, t)$, but replace the single-point Jacobian $J = \frac{\partial \hat{a}_1}{\partial a_t}$ with a Monte Carlo average over $K$ Gaussian
  perturbations of $a_t$:

  $$\hat{J}(a_t) = \frac{1}{K} \sum_{k=1}^{K} \frac{\partial \hat{a}_1}{\partial a_t}\Bigg|_{a_t + \epsilon_k}, \qquad \epsilon_k \overset{\text{i.i.d.}}{\sim} \mathcal{N}(0,,
  \sigma^2 I)$$

  The rest of the chain rule is unchanged. The full Q-gradient is:

  $$\nabla_{a_t} Q = \hat{J}(a_t)^\top  \nabla_{\hat{a}_1} Q(s, \hat{a}_1)$$

\paragraph{\methodname{} Regularized.}
We investigate an estimator where the Jacobian is regularized towards the identity:
\begin{equation}
    \label{eq:reguarlized-jacobian}
    \nabla_{a_t} Q(s, \hat{a}_1) = \nabla_{\hat{a}_1} Q(s, \hat{a}_1) \cdot (I + \epsilon \cdot \frac{\partial \hat{a}_1}{\partial a_t})
\end{equation}
where $\epsilon$ is some small value. We use $\epsilon=0.1$ in \cref{fig:qgf_reguarlized_jacobian}.

\paragraph{\methodname{}-Jacobian Ortho}
This is another regularization method for \methodname{}-Jacobian that orthogonalizes the Jacobian matrix. We orthogonalize with Singular Value Decomposition (SVD): $J = U \Sigma V^\top$, then 
$$J_{\text{ortho}} = U V^\top.$$
This is the nearest orthogonal matrix to $J$ (in Frobenius norm). The gradient then becomes:

$$\nabla_{a_t} Q(s, \hat{a}_1) = J_{\text{ortho}}^\top \nabla_{\hat{a}_1} Q(s, \hat{a}_1).$$

Intuitively, $J$ rotates and scales; orthogonalization keeps the rotation but discards the scaling, so the guidance direction is preserved but its magnitude is no longer amplified or shrunk by the singular values of the denoising map.

\cref{fig:qgf_reguarlized_jacobian} shows the offline RL performance of all the \methodname{} variants on OGBench.
Surprisingly, both \methodname{}-Jacobian Regularized and \methodname{}-Jacobian Ortho actually performs similarly to \methodname{}, and all three perform better than other variants with the Jacobian. The further shows that including the Jacobian actually hurts performance (likely due to the high variance of the gradient estimator). However, if one regularized the Jacobian, the performance can be recovered. That said, the simplest method remains just setting $J \approx \hat{J}=I$. Surprisingly, \methodname{} Distill actually also underperforms \methodname{}, and performs similarly to \methodname{}-chain. We hypothesize that it also suffers from the same constraint that \methodname{}-chain does, and places too much restrictions on strictly adhering to the dataset distribution.

\section{Additional \methodname{} Ablation Experiments}
\label{apdx:additional-ablations}

\begin{figure}[h]
    \centering
    \includegraphics[width=0.95\linewidth]{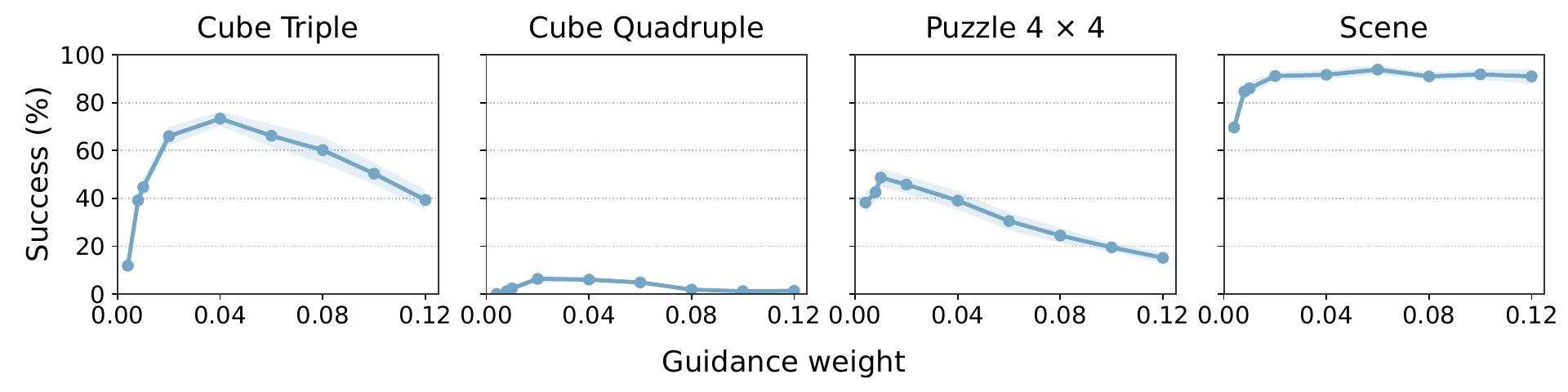}
    \caption{\textbf{\methodname{} performance against guidance weight}: increasing guidance weight can drastically improve policy performance, but overly large weight can also hurt performance by pushing actions off manifold.}
    \label{fig:guidance-weight-sensitivity}
\end{figure}

\paragraph{Guidance sensitivity analysis}.
We perform a sensitivity analysis of the guidance weight $\frac{1}{\beta}$ in \cref{eq:guidance_noise} for \methodname{} in \cref{fig:guidance-weight-sensitivity}. We can see that for most environments, increasing guidance weight drastically improves policy performance. However, an overly large guidance weight can also hurt performance since it can push the actions outside of the dataset support.

\begin{figure}[h]
    \centering
    \includegraphics[width=0.95\linewidth]{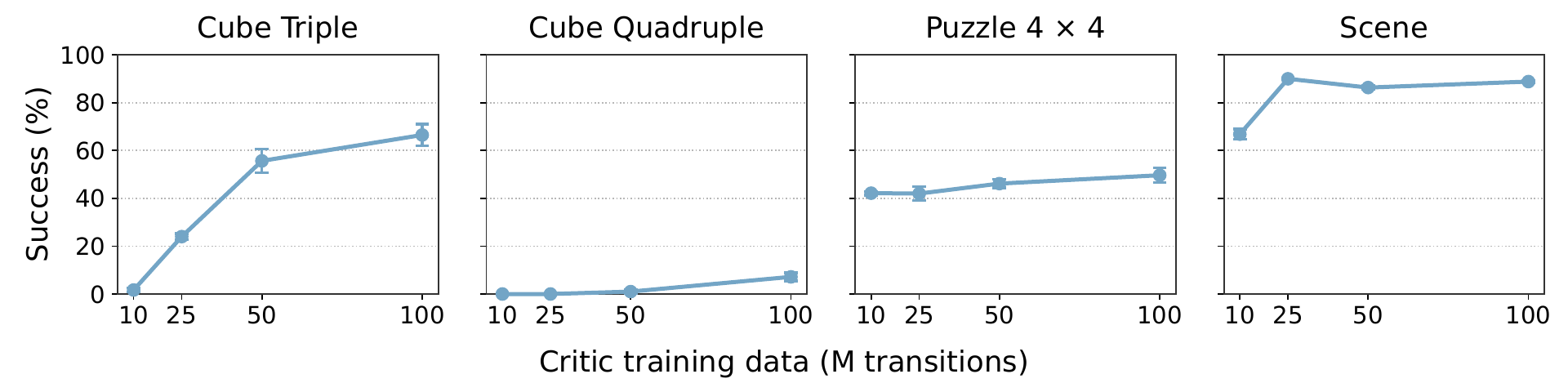}
    \caption{\textbf{Comparison of \methodname{}'s performance with critics of varying quality}, as induced by different dataset coverages, while the reference policy is fixed.}
    \label{fig:critic-quality}
\end{figure}

\paragraph{Critic sensitivity analysis.}
Here we analyze QGF's performance when subject to critics with varying quality, as induced by different offline dataset coverages. Specifically, we compare QGF's performance across critics trained on 100\%, 50\%, and 25\% of the offline dataset, while keeping the reference policy fixed (trained on the full offline dataset). A smaller subset of the offline dataset has less coverage, and therefore induces a less accurate critic. 
Results in \Cref{fig:critic-quality} show that critics trained on lower coverage data can hurt performance depending on the task. Intuitively, an inaccurate critic will give the wrong guidance signal regardless of whether the reference policy action is good, and guide action generation in wrong directions.

\begin{figure}[h]
    \centering
    \includegraphics[width=0.95\linewidth]{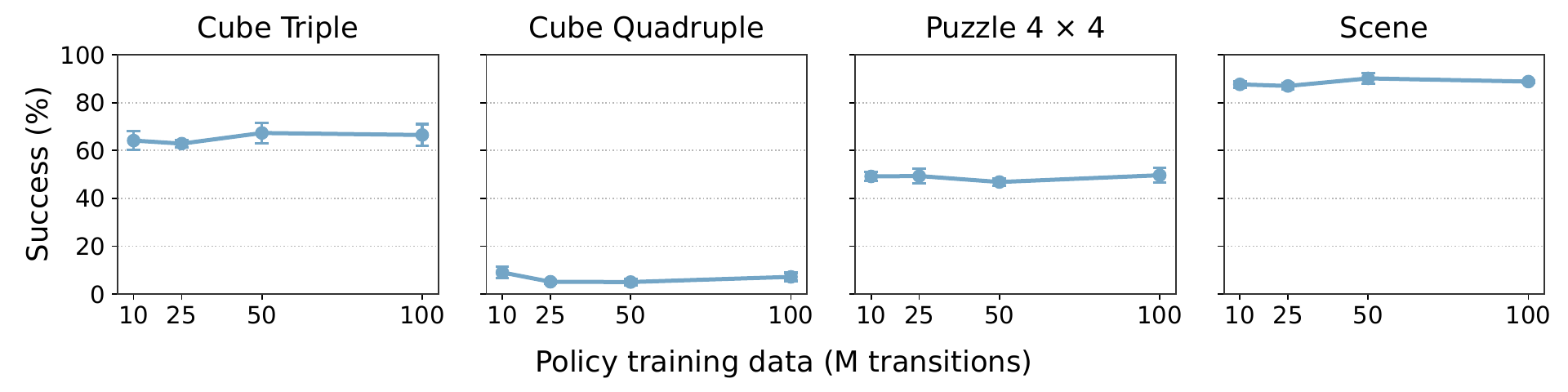}
    \caption{\textbf{Comparison of \methodname{}'s performance with reference BC policy of varying quality}, as induced by different dataset coverages, while the critic is fixed.}
    \label{fig:policy-quality}
\end{figure}

\paragraph{Reference policy sensitivity analysis.}
Similarly, we compare QGF's performance with different reference policy quality. Specifically, we compare QGF's performance using reference policies that are trained on 100\%, 50\%, and 25\% of the offline dataset, respectively, while keeping the critic fixed and trained on the full offline dataset. Results in \Cref{fig:policy-quality} shows that QGF has similar performance even when the reference policy has limited state-action coverage, showing that QGF can guide a weaker policy to achieve a similarly strong performance with Q gradient guidance.

\section{OOD Gradient Exploits $Q$ function}
\label{apdx:better-optimizer}

\begin{figure}[H]
    \centering
    \includegraphics[width=0.9\linewidth]{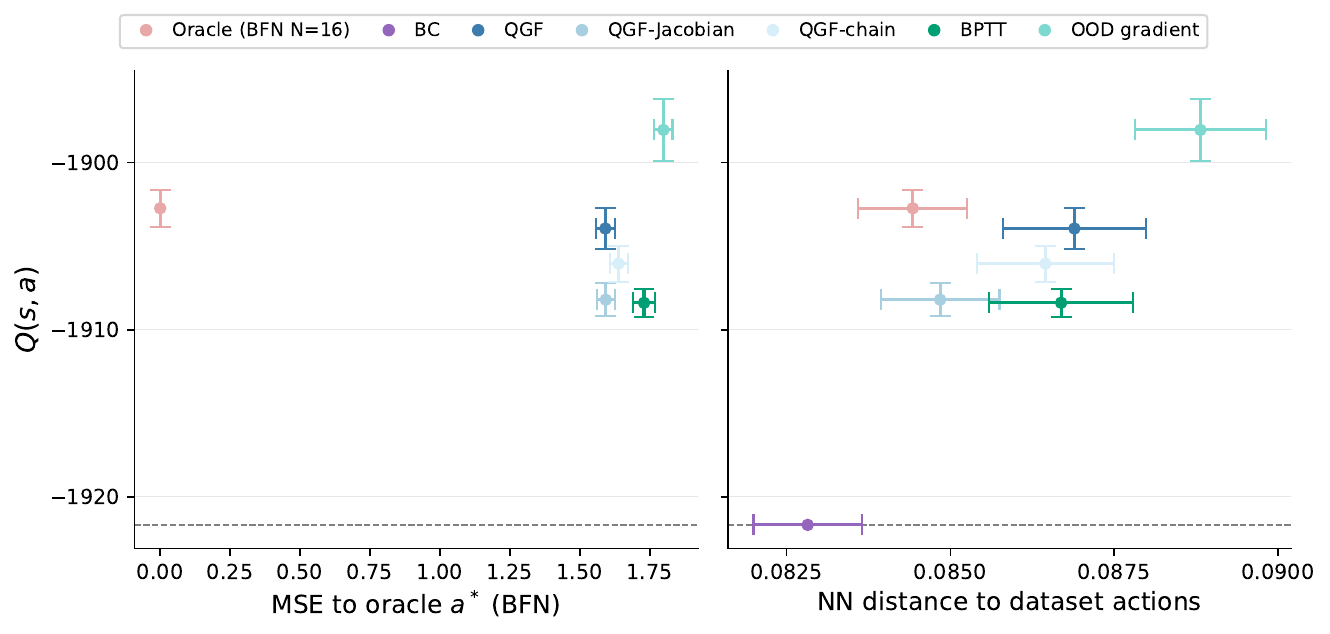}
    \caption{The OOD gradient leads to a denoised action with higher $Q$ value than all other methods by exploiting the critic on OOD actions. \textbf{(left)}: the OOD gradient leads to actions that are farthest away from the BFN oracle actions. \textbf{(right)}: the OOD gradient leads to actions that has the largest nearest-neighbor (NN) distance to dataset actions. Aggregated over $20$ tasks, $256$ observations, $4$ seeds.}
    \label{fig:q-vs-mse}
\end{figure}

\cref{fig:better-optimizer} shows that \methodname{} is an extremely effective optimizer for $Q$-values and does better than \methodname{}-Jacobian, \methodname{}-chain, and BPTT. It performs similarly to the oracle, which we take to be the best-of-n (BFN) sampling agent. The BFN agent is known to be extremely effective at optimizing actions to maximize $Q$-values~\citep{li2025reinforcement, hansen2023idql}, which aligns with our results in \cref{fig:result-bfn}.
Curiously, the OOD gradient is an even more effective optimizer than BFN, leading to even higher $Q$-values. However, \cref{fig:main-results} shows that the QFQL agent, which uses the OOD gradient, does not perform well in practice. This implies that the OOD gradient may be exploiting the $Q$ function by generating actions outside the distribution of the dataset (which we will call OOD actions) since the gradient itself is ill formed (see \cref{sec:rl-as-guidance}). This may not come as a surprise since it is well-known that critics trained with offline RL can over-estimate $Q$-values of OOD actions~\citep{kumar2020conservative, kostrikov2021offline, fujimoto2021minimalist}.
To investigate whether the actions from the OOD gradient are out of distribution, we plot both its distance to the oracle BFN action and its nearest-neighbor distance to actions from the offline dataset. \cref{fig:q-vs-mse} shows that actions from the OOD gradient has the largest distance on both distance metrics out of all the gradient estimators, showing that it produces actions least similar to dataset actions and optimal actions. This shows that the OOD gradient indeed denoised to some OOD actions that exploits the $Q$-values, and so such actions does not actually lead to good performance.

\section{\methodname{} for Offline-to-Online RL}
\label{apdx:online-rl}

\begin{figure}[h]
    \centering
    \includegraphics[width=\linewidth]{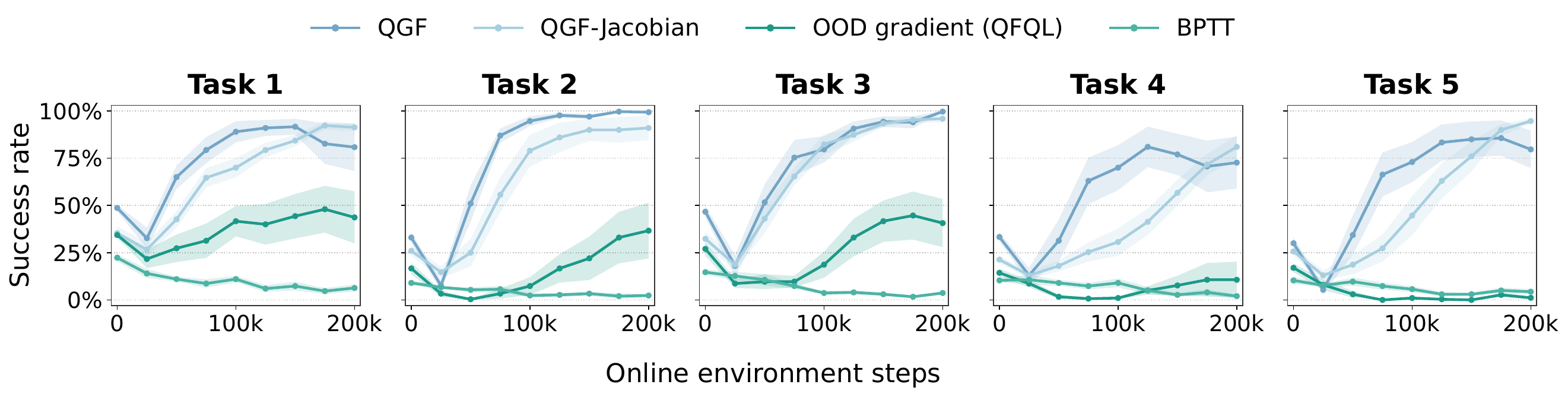}
    \caption{\textbf{Offline-to-online RL performance} (only online portion shown, 8 seeds) on \texttt{puzzle-4x4}: \methodname{} and \methodname{}-Jacobian performance significantly better than other test-time guidance methods.}
    \label{fig:offline-to-online}
\end{figure}

In \Cref{sec:results}, we mainly compare \methodname{} and baseline methods in the offline RL setting. Here, we explore whether test-time methods like \methodname{} is effective for online RL as well. Specifically, we assume the offline-to-online RL setting, where we initialize the policy and the value functions with offline pre-training and run online RL. During the online phase, we use $5000$ warmup steps~\citep{zhou2024efficient} where the offline-initializations are frozen for online data collection. After warmup, each online update samples $50/50$ from the offline dataset and the online replay buffer. We compare \methodname{} and \methodname{}-Jacobian against two representative test-time guidance methods: QFQL and BPTT.
\Cref{fig:offline-to-online} shows preliminary results on \texttt{puzzle-4x4}, where both \methodname{} and \methodname{}-Jacobian performs better than both baselines. Specifically, we observe that \methodname{} learns much faster than \methodname{}-Jacobian during the online phase, but \methodname{}-Jacobian eventually achieves a similar or slightly better performance given sufficient online samples. We hypothesize that this is another reflection of the the variance reduction effect of the \methodname{} gradient estimator: \methodname{}-Jacobian is less sample efficient because its gradient estimator has a higher-variance (see \Cref{fig:gradient-estimator-variance}), but eventually performs well because it is less biased.

\section{Critic Training Details}
\label{apdx:critic-training}
We follow the action chunking setting~\citep{li2025reinforcement} and learn both actors that produces action chunks and critics that are conditioned on action chunks (\ie $a_{1:h}$ for chunk size $h$).

\paragraph{Singletask IQL critic.}
Our method is agnostic to the specific critic training recipe used. In practice we chose to use IQL \cite{kostrikov2021offline} as the main critic training method because it can be trained using only in-sample learning and we can completely separate actor and critic training. All of the different methods in \cref{fig:main-results} are evaluated using an IQL Q function, which is learned as follows:

For an offline dataset $\mathcal{D}=\{(s_t,a_t,r_t,s_{t+1})\}$, we train a critic $Q_\theta$ and value function $V_\psi$ using the following objectives:
\begin{align}
\mathcal{L}_{Q}(\theta)
&=
\mathbb{E}_{(s_t,a_{t:t+h},r_{t:t+h},s_{t+h})\sim\mathcal{D}}
\Big[
\big(
Q_\theta(s_t,a_{t:t+h})-(\sum_{t'=t}^{t+h-1} [\gamma^{t'-t} r_t']+\gamma^h\,V_{\bar{\psi}}(s_{t+h})
\big)^2
\Big], \\
\mathcal{L}_{V}(\psi)
&=
\mathbb{E}_{(s_t,a_{t:t+h})\sim\mathcal{D}}
\Big[
L_2^\tau\!\big(
Q_{\bar{\theta}}(s_t,a_{t:t+h})-V_\psi(s_t)
\big)
\Big],
\end{align}
where $L_2^\tau(u)=|\tau-\mathbf{1}(u<0)|u^2$ is the expectile regression loss (expectile $\tau$), and $\bar{\theta},\bar{\psi}$ denote target/stop-gradient parameters. We train all IQL critics with an ensemble of $2$ critics and aggregate them by taking the minimum over the ensemble.

\paragraph{Goal-conditioned IQL critic.}
We train goal-conditioned IQL critics following the critic training procedure in Decoupled Q Chunking (DQC)~\citep{li2025decoupled}. DQC decouples the chunk horizon for the policy ($h_a$) and the critic ($h_c$). The chunk horizon used for the critic is usually much larger to speed up value propagation. We train a total of $2$ critics, $Q_c$ and $Q_a$, which has chunk horizon $h_c$ and $h_a$ respectively. 

The long-horizon critic $Q_c$ is trained using the same IQL objective as above:
\begin{align}
\mathcal{L}{Q_c}(\theta_c)
&=
\mathbb{E}_{(s_t,a_{t:t+h_c},r_{t:t+h_c},s_{t+h_c})\sim\mathcal{D}}
\Big[
\big(
Q_c(s_t,g,a_{t:t+h_c}) \
 -
(
\sum_{t’=t}^{t+h_c-1}\gamma^{t’-t} r_{t’}
+\gamma^{h_c}V_{\bar{\psi}}(s_{t+h_c},g)
)
\big)^2
\Big],
\end{align}
with the corresponding expectile value objective
\begin{align}
\mathcal{L}_{V}(\psi)
\mathbb{E}_{(s_t,g,a_{t:t+h_c})\sim\mathcal{D}}
\Big[
L_2^\tau\!\big(
Q_{\bar{\theta}c}(s_t,g,a_{t:t+h_c})
V_\psi(s_t,g)
\big)
\Big].
\end{align}

The short-horizon critic $Q_a$ is trained by distilling the optimistic value of extending a partial action chunk into a complete chunk. Concretely, for a partial chunk $a_{t:t+h_a}$, we regress $Q_a$ towards an upper expectile of the full-chunk critic values:
\begin{align}
\mathcal{L}_{Q_a}(\theta_a)
\mathbb{E}_{(s_t,g,a_{t:t+h_c})\sim\mathcal{D}}
\Big[
L_2^{\kappa_d}
\!\Big(
Q_c(s_t,g,a_{t:t+h_c}) -
Q_a(s_t,g,a_{t:t+h_a})
\Big)
\Big],
\end{align}
where $\kappa_d > 0.5$ is the distillation expectile. This objective causes $Q_a$ to approximate the value of the best completion of the partial chunk under $Q_c$:
\begin{align}
Q_a(s_t,g,a_{t:t+h_a})
\approx
\max_{a_{t+h_a:t+h_c}}
Q_c(s_t,g,[a_{t:t+h_a},a_{t+h_a:t+h_c}]).
\end{align}

For policy extraction, we optimize and evaluate policies using only $Q_a$. This allows the policy to operate with a short execution horizon $h_a$ while still benefiting from the faster value propagation provided by the long-horizon critic $Q_c$.

\section{\methodname{} with Other Types of Value Functions}
\label{apdx:other-value-functions}

\begin{figure}[th]
    \centering
    \includegraphics[width=\linewidth]{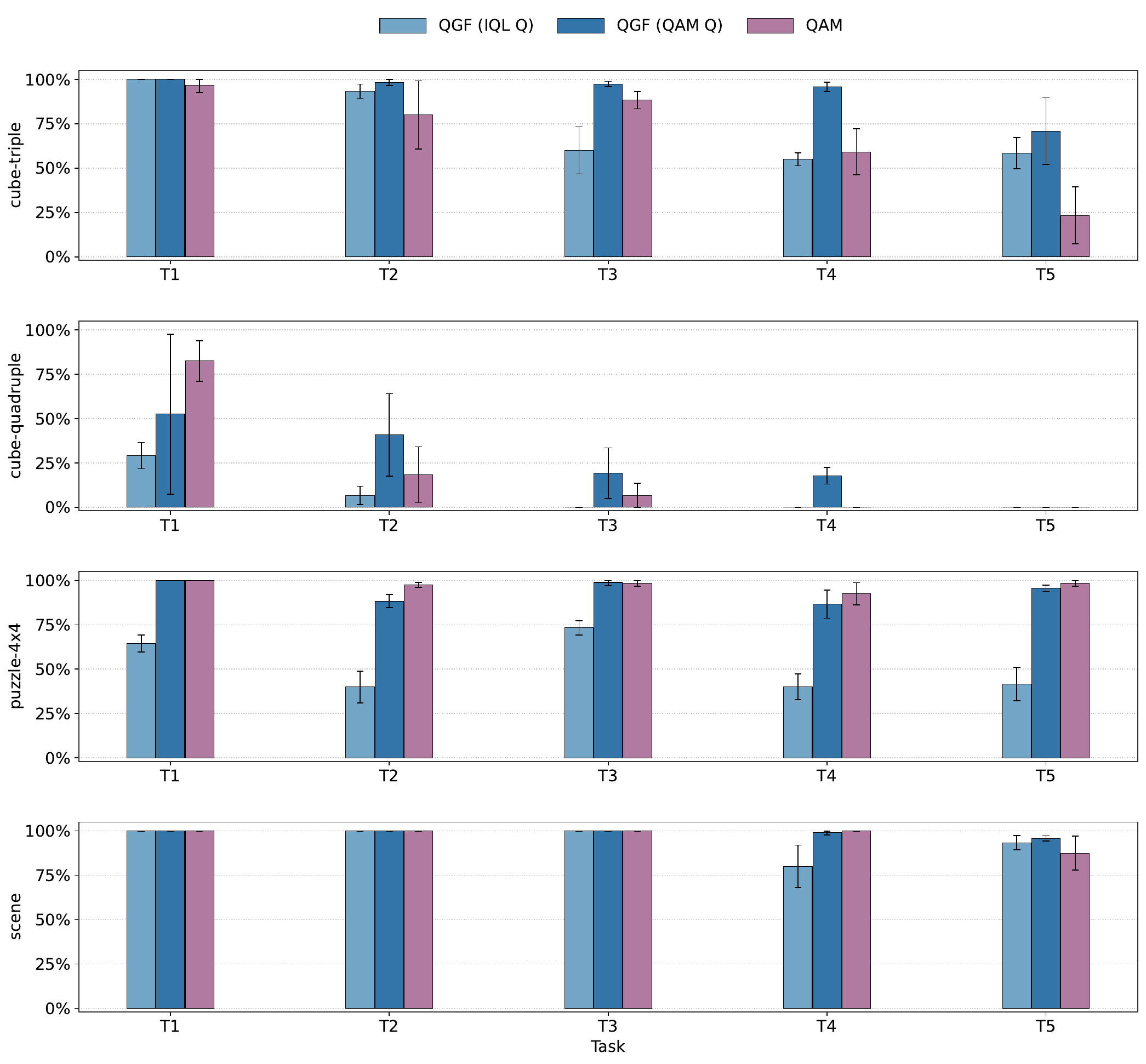}
    \caption{Task breakdown of \cref{fig:qam-value-functions}: \methodname{} with a QAM-based critic performs even better than \methodname{} with an IQL-based critic on OGBench.}
    \label{fig:qam-value-functions-full}
\end{figure}

In \cref{sec:different-critics}, we train the critic by minimizing the temporal difference (TD) error:
$$\mathbb{E}_{s, a, s' \in \mathcal{D}}\left(Q(s,a) - r(s,a) - \gamma \mathbb{E}_{a' \sim \pi_{\text{QAM}}}[Q_{\bar \phi}(s', a')]\right)^2.$$
To get the best critic possible, we following the original QAM paper~\citep{li2026q} and train an ensemble of $10$ $Q$ functions, and aggregate them by taking the mean of the ensemble values.
The full result is presented in \cref{fig:qam-value-functions-full}. To avoid confusion, the QAM agent in \cref{fig:qam-value-functions} and \cref{fig:qam-value-functions-full} is trained using the critic described above by minimizing TD error, while the QAM agent in \cref{fig:main-results} uses IQL critic for comparability against other agents.

\section{Details On Baseline Methods}
\label{apdx:baselines-description}
\paragraph{(1) Test-time methods}

\paragraph{Best-of-N (BFN)} performs rejection sampling on the base behavior cloning policy $\hat\pi$ trained with the flow matching objective. Specifically, the output action $a$ is selected as: 
\begin{equation}
    a \leftarrow \argmax_{a_1, \cdots,a_N \sim \hat{\pi}}Q(s,a_i).
\end{equation} 

\paragraph{BPTT} is a baseline we consider inspired by DQL \cite{wang2022diffusion} that computes guidance by backpropagating the critic objective through the \emph{entire} denoising chain (i.e., full Backpropogation through time), yielding $\nabla_{a_t} Q(s, a_1)$ where $a_1$ is the action produced after all denoising steps, and $a_t$ is the partially-denoised action at timestep $t$.
We compute the denoised action as 
$$a_1 = \mathrm{ODE}(a_t) = D_\theta^{(1)} \circ \cdots \circ D_\theta^{(t+\delta)} \circ D_\theta^{(t)}(s, a_t),$$
where $D_\theta^{(t)}$ denotes the $t$-th denoising step of the policy.
Then we can write the gradient as:
\begin{equation}
    \nabla_{a_t} Q(s,\, D_\theta^{(1)} \circ \cdots \circ D_\theta^{(t+\delta)} \circ D_\theta^{(t)}(s, a_t)),
\end{equation}
  and taking the gradient at $a_t$ requires differentiating through the chain of denoising steps $D_\theta^{(t)} \rightarrow \cdots \rightarrow D_\theta^{(1)}.$

\paragraph{GradStep} is a baseline we consider inspired by Policy Agnostic RL (PA-RL) \cite{mark2024policy} that first samples a denoised action by running the full denoising process of our behavior cloning flow policy $\hat\pi$, then performs policy improvement by iteratively taking $L$ gradient ascent steps on the fully denoised action:

\begin{algorithm}[h]
\caption{GradStep}
\centering
\begin{algorithmic}
\State \textbf{Input:} $L, \alpha_{\mathrm{step}}, \hat{\pi}$,
\State $a \leftarrow a_0$, where $a_0 \sim \hat\pi$
\For{$\ell = 1,\ldots,L$}
    \State $a \leftarrow a + \alpha_{\mathrm{step}} \nabla_a Q_\theta(s,a)$
\EndFor
\State \Return $a$
\end{algorithmic}
\end{algorithm}

\paragraph{QFQL \cite{jang2025q}} also trains a behavior cloning actor via flow matching and critic using TD Learning, and perform policy optimization at test time with $Q$ critic gradeint guidance.
QFQL uses the critic gradient at the intermediate noisy action $a_t$ directly (i.e., $\nabla_{a_t} Q(s, a_t)$), which is out-of-distribution  for a critic trained only on clean actions. During inference, the denoising step at time $t$ is characterized by:
\begin{equation}
    a_{t+\delta} \leftarrow a_t + \delta \left( v_\theta(s, a_t, t) + \frac{1}{\beta} \nabla_{a_t} Q(s, a_t) \right)
\end{equation}

\paragraph{RobustQ} is a novel baseline we consider, inspired by the adversarially robust classifiers~\citep{santurkar2019image, kawar2022enhancing, tsipras2018robustness, ilyas2019adversarial} in the text-to-image generation classifier guidance literature. To handle querying $Q$ gradient at the noisy action step $a_t$, we train a $Q$ function $Q(s, a_t, t)$ to regress on the IQL $Q$ with loss:
\begin{equation}
    l_{robust} = (Q(s, a_t, t) - Q(s, a))^2.
\end{equation}
Then, we query the gradient $\nabla_{a_t} Q(s, a_t, t)$ during inference time and use that as guidance.

\paragraph{CFGRL \cite{frans2025diffusion}} also performs policy improvement during test time, but does not rely on critic gradient. During training, CFGRL trains a base flow matching policy $f(s, a_t, t)$ and an optimality-conditioned policy $f_{o=1}(s, a_t,t)$. $o$ is a binary indicator on whether the policy is ``optimal'', and $o=1$ means it is. Following the original paper, we define ``optimal'' as $0 < A(s, a) = Q(s, a) - V(s)$. The base policy $f$ maps Gaussian noise to the whole dataset distribution, while the conditioned policy $f_{o=1}$ maps noise to the subset of the dataset distribution that is ``optimal''. 
During inference, CFGRL performs classifier free guidance (CFG) to combine the two policies into a single guided policy:
$$v_{\mathrm{cfg}}(s, a_t, t) = f(s, a_t, t) + w (f_{o=1}(s, a_t, t) - f(s, a_t, t)),$$
where $w$ is the CFG weight that controls the balance between the two velocities.

\paragraph{(2) Training-time methods}

\paragraph{Q-Score Matching (QSM) \cite{psenka2023learning}} trains a diffusion policy to 
follow the Boltzmann distribution of the Q-function, $\pi(\cdot \mid s) \propto 
e^{Q(s, \cdot)/\beta}$ by approximating the score of intermediate noised actions 
$a_t = \sqrt{\alpha_t} a + \sqrt{1 - \alpha_t} \varepsilon$ directly with the action 
gradient of the critic:
\begin{equation}
    \mathcal{L}_{\text{QSM}}(\theta) = \mathbb{E}_{\varepsilon \sim \mathcal{N},\, 
    t \sim \mathcal{U}\{0, \frac{1}{T}\ldots,1 - \frac{1}{T}\}} \left[ \left\| {-\frac{1}{\beta} \nabla_{a_t} Q(s, a_t) 
    - f_\theta(s, a_t, t)} \right\|_2^2 \right].
\end{equation}
Note that QSM essentially hill climbs with the OOD gradient direction (\ie $\nabla_{a_t} Q(s, a_t)$) during training.
In practice, we find that QSM alone often does not perform well in offline RL settings. Therefore, we try to improve its performance by augmenting the QSM objective with a standard
DDPM behavior regularization term $\mathcal{L}_{\text{DDPM}}(\theta) = \mathbb{E}_{\varepsilon \sim \mathcal{N},\, t \sim \mathcal{U}\{0,\frac{1}{T}\ldots,1-\frac{1}{T}\}} \left[ \left\| \varepsilon - f_\theta(s, a_t, i) \right\|_2^2 \right]$. We call this agent QSM-BC. QSM-BC has the overall 
actor loss $\mathcal{L}_{\mathrm{QSM-BC}}(\theta) = \mathcal{L}_{\mathrm{QSM}}(\theta) + \eta 
\mathcal{L}_{\mathrm{DDPM}}(\theta)$. 

\paragraph{Diffusion Actor-Critic (DAC) \cite{fang2024diffusion}} is similar to QSM, but 
incorporates a behavior prior constraint: it aims to learn policy $\pi(\cdot |s) \propto \hat{\pi}(\cdot|s) e^{Q(s, \cdot)}$. Therefore, they assume that
$$\nabla_{a_t} \log \pi(a_t \mid s) \approx \nabla_{a_t} \log \hat{\pi}(a_t \mid s) + \frac{1}{\beta} 
\nabla_{a_t} Q(s, a_t).$$
The actor is trained by matching $f_\theta$ to a linear 
combination of the behavior cloning target $\varepsilon$ and the Q-gradient:
\begin{equation}
    \mathcal{L}_{\text{DAC}}(\theta) = \mathbb{E}_{\varepsilon \sim \mathcal{N},\, 
    t \sim \mathcal{U}\{0,\ldots,T-1\}} \left[ \left\| \varepsilon - \frac{1}{\beta} \nabla_{a_t} 
    Q(s, a_t) - f_\theta(s, a_t, t) \right\|_2^2 \right].
\end{equation}
DAC can be interpreted as a training-time version of QFQL, which ascends on the OOD gradient direction with a behavior constraint.

\paragraph{Flow Q-Learning (FQL) \cite{park2025flow}} avoids the issue of backpropagating through the denoising chain by 
distilling a flow matching behavior cloning policy $f_\theta : \mathcal{S} \times 
\mathbb{R}^A \times [0,1] \rightarrow \mathbb{R}^A$ into a one-step policy 
$f_\omega : \mathcal{S} \times \mathbb{R}^A \rightarrow \mathbb{R}^A$. 
$f_\theta$ is trained with a standard flow matching objective, and the one-step policy objective is:
\begin{equation}
    \mathcal{L}_{\mathrm{onestep}}(\omega) = \mathbb{E}_{z \sim \mathcal{N}} \left[ 
    \alpha_{\mathrm{FQL}} \left\| f_\omega(s, z) - \mathrm{ODE}(f_\theta(s, \cdot, \cdot), z) \right\|_2^2 
    - Q\!\left(s, f_\omega(s, z)\right) \right],
\end{equation}
where $\alpha_{\mathrm{FQL}}$ is a coefficient balancing proximity (in 2-Wasserstein distance) to the behavior policy 
$\pi_{\theta}$ with $Q$-value maximization. 

\paragraph{Q-learning with Adjoint Matching (QAM) \cite{li2026q}} addresses the 
instability of BPTT by leveraging adjoint matching \cite{domingoenrich2025adjointmatchingfinetuningflow} to 
directly incorporate the critic's action gradient into the flow policy training without 
backpropagating through the denoising process. Specifically, QAM targets the optimal 
behavior-constrained policy $\pi(a \mid s) \propto \hat{\pi}(a \mid s) 
e^{Q(s,a)/\beta}$ by solving the following stochastic optimal control (SOC) objective 
via the adjoint matching loss:
\begin{equation}
    \mathcal{L}_{\text{AM}}(\theta) = \mathbb{E}_{s, \{a_t\}_t} \left[ \int_0^1 
    \left\| \frac{2(f_\theta(s, a_t, t) - \hat{f}(s, a_t, t))}{\sigma_t} + 
    \sigma_t \tilde{g}_t \right\|_2^2 dt \right],
\end{equation}
where $\tilde{g}_t$ is the ``lean'' adjoint state propagated backwards through the 
base flow model $\hat{f}$ via
\begin{equation}
    d\tilde{g}_t = -\nabla_{a_t}\left[2\hat{f}(s, a_t, t) - a_t/t\right] 
    \tilde{g}_t \, dt, \quad \tilde{g}_1 = - \nabla_{a_1} Q(s, a_1) / \beta.
\end{equation}
Unlike approximation-based methods (e.g., QSM, DAC), the adjoint state computation 
uses only $\hat{f}$ rather than $f_\theta$, avoiding backpropagating through the policy that is being optimized.

\paragraph{Efficient Diffusion Policy (EDP)~\citep{kang2023efficient}} avoid the full backpropagation through time by using a one-Euler-step approximation of the full denoising process. Specifically, given $a_t=(1-t)\epsilon+ta$, EDP approximates the full denoising process with $\hat{a}_1(a_t)=a_t+(1-t)v_\theta(a_t,s,t)$. The policy is trained to maximize $Q$-values subject to a behavior constraint with loss:
\begin{equation}
    \mathcal{L}_{\mathrm{EDP}}(\theta)
    =
    \mathbb{E}\left[
        -Q(s,\hat{a}_1(a_t))
        + \alpha\left\|v_\theta(a_t,s,t)-(a-\epsilon)\right\|_2^2
    \right].
\end{equation}
During training, the action gradient passes through the one-step action approximation and is taken with respect to $a_t$. The coefficient $\alpha$ controls the strength of behavior regularization.

\section{Limitations}
\label{apdx:limitation}

Here we discuss the limitations of our proposed RL algorithm \methodname{}. As discussed, our method relies on using the critic gradient to guide action optimization at test time. While this is generally much cheaper than other test-time RL algorithms such as best-of-N sampling, a critic parameterized by a large model may be expensive to take gradients over. Further work is needed in this direction to optimize the inference compute requirement of our method for large critics.
Second, \methodname{} relies on adjusting the output of a base reference model. If the base model is under-trained or does not well represent the data distribution, using \methodname{} to improve at test time would not be efficient. One interesting direction of future research is better training recipe for reference models for \methodname{}. Third, we assume the setting where rewards are available in the offline dataset.

\section{Hyperparameter and Tuning Details}
\label{apdx:hyperparams}
For all the main results in \cref{fig:main-results}, \cref{fig:result-bfn}, and \cref{fig:dqc-results}, we run $10$ different seeds. All other ablations experiment include $4$ seeds or more. All performance result plots in the paper report 95\% confidence interval as the error bar. 

\cref{tab:hyperparams} below denote the common hyperparameters and \cref{tab:domain_hyperparams} denote the method-specific hyperparameters. For each method, including \methodname{} and all baseline methods, we tune the method-specific hyperparameter per domain in the ranges specified in \cref{tab:sweep_ranges_100m_chunking}. Specifically, we tune hyperparameters per domain on task 2 and 4 only, and re-run all $10$ seeds on task 1 through 5. We choose this hyperparameter tuning since task difficulty gradually increases from task 1 to task 5, and so tuning on task 2 and 4 represent the average performance on the domain.

\label{apdx:hyperparam}
\begin{table}[H]
\centering
\begin{tabular}{cc}
\toprule
\textbf{Parameter} & \textbf{Value} \\
\midrule
Batch size & 1024 \\
Discount factor ($\gamma$) & 0.999 \\
IQL Critic Expectile ($\tau$) & 0.9 \\
Action chunk horizon ($h$) & 5 \\
Flow steps & 10 \\
Offline training steps & $5 \times 10^{5}$ \\
Evaluation episodes & 10 \\
Critic Network Size & [1024, 1024, 1024, 1024] \\
Actor Network Size & [1024, 1024, 1024, 1024] \\
Learning rate & 3e-4 \\
Critic ensemble size & 2 \\
Critic aggregation method & min \\
\bottomrule
\end{tabular}
\vspace{0.3em}
\caption{\textbf{Common hyperparameters} used for singletask Offline RL experiments on OGBench $100$M datasets.}
\label{tab:hyperparams}
\end{table}

\label{apdx:hyperparam}
\begin{table}[H]
\centering
\begin{tabular}{cc}
\toprule
\textbf{Parameter} & \textbf{Value} \\
\midrule
Batch size & 4096 \\
Discount factor ($\gamma$) & 0.999 \\
Flow steps & 10 \\
Offline training steps & $1 \times 10^{6}$ \\
Evaluation episodes & 50 per task \\
Critic Network Size & [1024, 1024, 1024, 1024] \\
Actor Network Size & [1024, 1024, 1024, 1024] \\
Learning rate & 3e-4 \\
Critic ensemble size & 2 \\
Critic aggregation method & mean for puzzle-*, min for cube-* \\
\bottomrule
\end{tabular}
\vspace{0.3em}
\caption{\textbf{Common hyperparameters} used for offline goal-conditioned RL experiments on OGBench.}
\label{tab:hyperparams}
\end{table}

\begin{table}[H]
\centering
\small
\setlength{\tabcolsep}{2.9pt}
\begin{tabular}{lccccccccccccc}
\toprule
& \multicolumn{5}{c}{\textit{Training-time}}
& \multicolumn{8}{c}{\textit{Test-time}} \\
\cmidrule(lr){2-6} \cmidrule(lr){7-14}
Domains
& DAC
& \shortstack{QSM\\+BC}
& FQL
& QAM
& EDP
& \shortstack{Grad\\Step}
& CFGRL
& BPTT
& QFQL
& \shortstack{Robust\\Q}
& \shortstack{\methodname{}\\Jac.}
& \textbf{\methodname{}}
& \shortstack{\textbf{\methodname{}}\\\textbf{BFN}} \\
&
$(\alpha)$
& $(\tau_{\mathrm{inv}}, \alpha)$
& $(\alpha)$
& $(\tau_{\mathrm{inv}})$
& $(w_{\mathrm{BC}})$
& $(\alpha_{\mathrm{step}}, L)$
& $(\tau_{\mathrm{cfg}})$
& $(\tau_{\mathrm{g}})$
& $(\tau_{\mathrm{g}})$
& $(\tau_{\mathrm{g}})$
& $(\tau_{\mathrm{g}})$
& $(\tau_{\mathrm{g}})$
& $(\tau_{\mathrm{g}})$ \\
\midrule
cube-triple-100M-*    & 100 & (0.1, 10) & 1000 & 0.1 & 100 & (0.01, 3) & 3.0 & 0.06 & 0.06 & 0.12 & 0.04 & 0.06 & 0.02 \\
cube-quadruple-100M-* & 100 & (0.1, 10) & 1000 & 0.1 & 300 & (0.01, 3) & 3.0 & 0.06 & 0.06 & 0.10 & 0.02 & 0.10 & 0.008 \\
puzzle-4x4-100M-*     & 100 & (0.1, 10) & 1000 & 0.1 & 300 & (0.01, 3) & 2.0 & 0.008 & 0.02 & 0.10 & 0.008 & 0.02 & 0.004 \\
scene-100M-*          & 300 & (0.1, 10) & 300  & 0.1 & 30  & (0.01, 5) & 5.0 & 0.12 & 0.04 & 0.12 & 0.008 & 0.12 & 0.02 \\
\bottomrule
\end{tabular}
\caption{\textbf{Domain-specific hyperparameters} for single-task OGBench experiments.}
\label{tab:domain_hyperparams}
\end{table}

\begin{table}[H]
\centering
\resizebox{0.85\textwidth}{!}{%
\begin{tabular}{lcccccccc}
\toprule
 & \multicolumn{4}{c}{Guidance weight ($\tau_g$)} & \multicolumn{4}{c}{DQC critic} \\
\cmidrule(lr){2-5} \cmidrule(lr){6-9}
Domains & BPTT & QFQL & QGF-Jacobian & QGF & $h_c$ & $h_a$ & $\tau$ & $\kappa_d$ \\
\midrule
puzzle-4x5-3M     & 10 & 10 & 3  & 3  & 25 & 5 & 0.9  & 0.5 \\
cube-triple-100M    & 3  & 3  & 3  & 3  & 25 & 5 & 0.93 & 0.8 \\
cube-quadruple-100M & 10 & 3  & 10 & 3  & 25 & 5 & 0.93 & 0.8 \\
puzzle-4x6-1B     & 1  & 1  & 1  & 1  & 25 & 1 & 0.7  & 0.5 \\
cube-octuple-1B   & 10 & 10 & 10 & 10 & 25 & 5 & 0.93 & 0.5 \\
\bottomrule
\end{tabular}%
}
\caption{\textbf{Domain-specific guidance weights and DQC value function hyperparameters} for goal-conditioned RL experiments on OGBench. $h_c$ and $h_a$ are the critic and policy action chunk horizons; $\kappa_b$ and $\kappa_d$ are the backup and distillation expectiles used by the Decoupled Q-Chunking (DQC) critic. These are tuned values taken from the DQC paper.}
\label{tab:domain_hyperparams_dqc}
\end{table}

\begin{table}[H]
\centering
\small
\setlength{\tabcolsep}{8pt}
\renewcommand{\arraystretch}{1.2}
\begin{tabularx}{\linewidth}{l c >{\raggedright\arraybackslash}X}
\toprule
\textbf{Method} & \textbf{Hyperparameter(s)} & \textbf{Tuning Range} \\
\midrule
\multicolumn{3}{l}{\textit{Training-time methods}} \\
DAC       & $\alpha$                       & $\{100, 300, 1000\}$ \\
QSM+BC    & $(\tau_{\mathrm{inv}}, \alpha)$ & $\{0.1, 3.0\} \times \{10, 30\}$ \\
FQL       & $\alpha$                       & $\{10, 100, 300, 1000\}$ \\
QAM (IQL) & $\tau_{\mathrm{inv}}$           & $\{0.1, 0.3, 1.0\}$ \\
EDP       & $w_{\mathrm{BC}}$               & $\{1, 10, 30, 100, 300\}$ \\
CFGRL     & $\tau_{\mathrm{cfg}}$           & $\{1.0, 2.0, 3.0, 5.0, 10.0, 15.0\}$ \\
\midrule
\multicolumn{3}{l}{\textit{Test-time methods}} \\
GradStep  & $(\eta, K)$                    & $\{0.01, 0.03, 0.1\} \times \{1, 3, 5\}$ \\
QGF       & $\tau_g$                       & $\{0.004, 0.008, 0.01, 0.02, 0.04, 0.06, 0.08, 0.1, 0.12\}$ \\
QFQL      & $\tau_g$                       & $\{0.004, 0.008, 0.01, 0.02, 0.04, 0.06, 0.08, 0.1, 0.12\}$ \\
BPTT      & $\tau_g$                       & $\{0.004, 0.008, 0.01, 0.02, 0.04, 0.06, 0.08, 0.1, 0.12\}$ \\
BFN       & $N$                            & $\{4, 8, 16, 32\}$ \\
QGF+BFN   & $(N, \tau_g)$                  & $\{4, 8, 16, 32\} \times \{0.004, 0.008, 0.01, 0.02, 0.04, 0.06, 0.08, 0.1, 0.12\}$ \\
\bottomrule
\end{tabularx}
\caption{\textbf{Hyperparameter tuning ranges} for OGBench experiments.}
\label{tab:sweep_ranges_100m_chunking}
\end{table}

\section{Detailed Results}
\label{apdx:result-details}

\begin{figure}[H]
    \centering
    \includegraphics[width=1\linewidth]{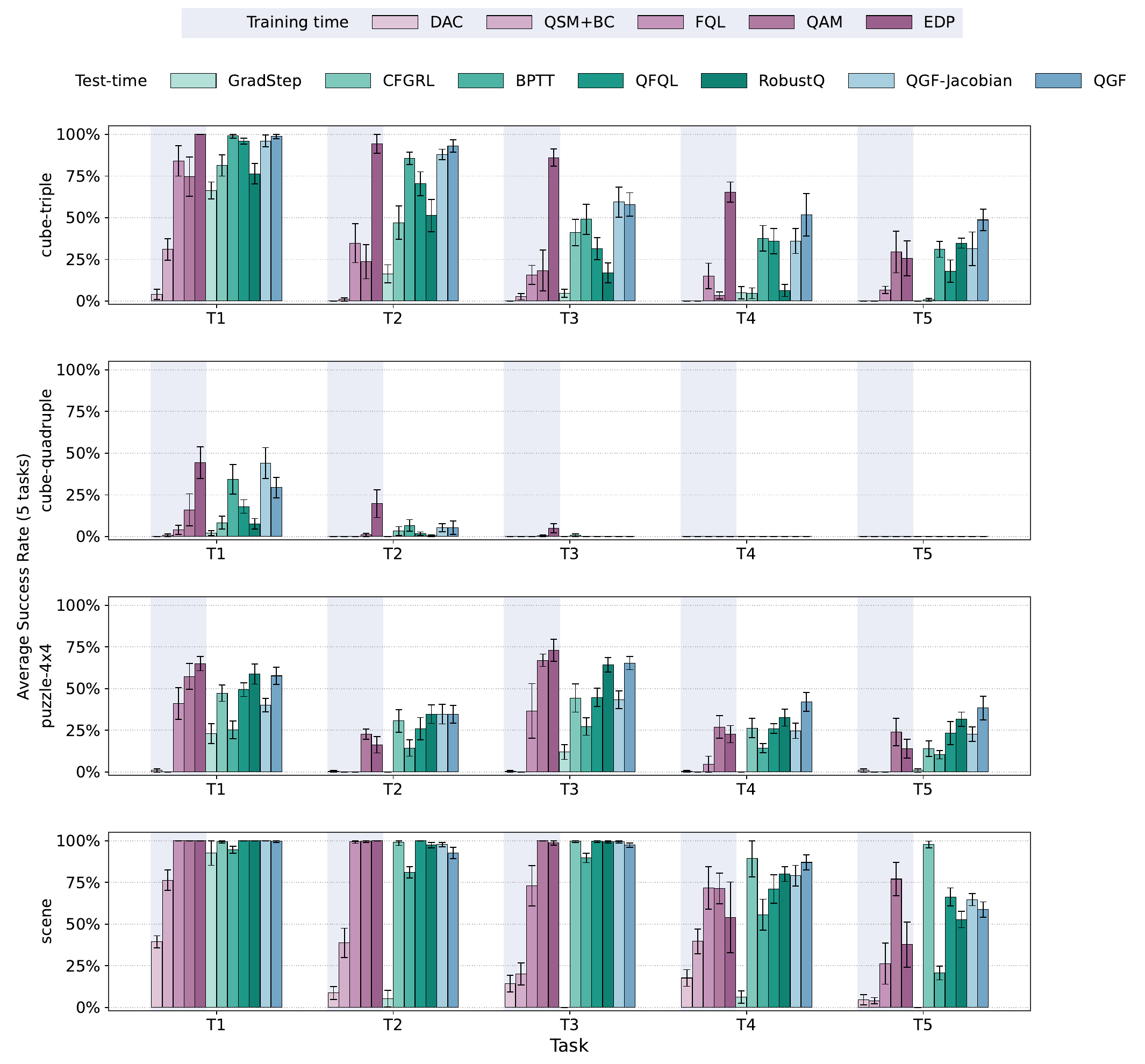}
    \vspace{-0.5cm}
    \caption{Task breakdown of \Cref{fig:main-results}: offline RL performance at 500k training steps per environment ($10$ seeds each). See the exact numbers in \Cref{tab:main-results}}
    \label{fig:full-result-no-bfn}
\end{figure}

\begin{table}[h]
\definecolor{qgfblue}{HTML}{0076BA}
\definecolor{qgflightblue}{HTML}{EDF5FA}
\definecolor{qgftrainpurple}{HTML}{9B5F8E}
\definecolor{qgftestgreen}{HTML}{108274}
\def\qgfval#1{\cellcolor{qgflightblue}\textcolor{qgfblue}{\ensuremath{#1}}}
\def\qgfjacval#1{\qgfval{#1}}
\def\qgftrainval#1{\cellcolor{qgftrainpurple!10}\textcolor{qgftrainpurple}{\ensuremath{#1}}}
\def\qgftestval#1{\cellcolor{qgftestgreen!10}\textcolor{qgftestgreen}{\ensuremath{#1}}}
    \centering
    \normalfont\mathversion{normal}
    \captionsetup{format=plain,font=normalfont,labelfont=normalfont,textfont=normalfont}
    \small
    \setlength{\tabcolsep}{2pt}
    \renewcommand{\arraystretch}{1.08}
    \resizebox{\linewidth}{!}{%
    \begin{tabular}{@{}llrrrrrrrrrrrr@{}}
        \toprule
        Env. & Task & \multicolumn{1}{c}{\textcolor{qgftrainpurple}{DAC}} & \multicolumn{1}{c}{\textcolor{qgftrainpurple}{QSM+BC}} & \multicolumn{1}{c}{\textcolor{qgftrainpurple}{FQL}} & \multicolumn{1}{c}{\textcolor{qgftrainpurple}{QAM}} & \multicolumn{1}{c}{\textcolor{qgftrainpurple}{EDP}} & \multicolumn{1}{c}{\textcolor{qgftestgreen}{GradStep}} & \multicolumn{1}{c}{\textcolor{qgftestgreen}{CFGRL}} & \multicolumn{1}{c}{\textcolor{qgftestgreen}{BPTT}} & \multicolumn{1}{c}{\textcolor{qgftestgreen}{QFQL}} & \multicolumn{1}{c}{\textcolor{qgftestgreen}{RobustQ}} & \multicolumn{1}{c}{\textcolor{qgfblue}{\shortstack{QGF-\\Jacobian}}} & \multicolumn{1}{c}{\textcolor{qgfblue}{QGF}} \\
        \midrule
         & T1  & \qgftrainval{4.0 \pm 3.0} & \qgftrainval{31.0 \pm 6.5} & \qgftrainval{84.0 \pm 9.1} & \qgftrainval{74.7 \pm 11.8} & \qgftrainval{100.0 \pm 0.0} & \qgftestval{66.3 \pm 5.1} & \qgftestval{81.3 \pm 6.3} & \qgftestval{99.3 \pm 0.8} & \qgftestval{96.0 \pm 2.6} & \qgftestval{76.3 \pm 6.2} & \qgfjacval{97.0 \pm 3.0} & \qgfval{98.7 \pm 1.9} \\
         & T2  & \qgftrainval{0.0 \pm 0.0} & \qgftrainval{1.0 \pm 1.0} & \qgftrainval{34.7 \pm 11.7} & \qgftrainval{23.7 \pm 10.3} & \qgftrainval{94.3 \pm 5.8} & \qgftestval{16.3 \pm 5.4} & \qgftestval{47.0 \pm 10.0} & \qgftestval{85.7 \pm 3.7} & \qgftestval{70.3 \pm 7.2} & \qgftestval{51.3 \pm 9.7} & \qgfjacval{88.0 \pm 3.1} & \qgfval{93.0 \pm 3.7} \\
        \raisebox{-\baselineskip}[0pt][0pt]{\shortstack[l]{Cube\\Triple}} & T3  & \qgftrainval{0.0 \pm 0.0} & \qgftrainval{2.7 \pm 1.9} & \qgftrainval{15.7 \pm 5.8} & \qgftrainval{18.3 \pm 12.3} & \qgftrainval{86.0 \pm 5.1} & \qgftestval{4.7 \pm 2.4} & \qgftestval{41.0 \pm 7.9} & \qgftestval{49.0 \pm 9.1} & \qgftestval{31.3 \pm 6.6} & \qgftestval{17.0 \pm 5.9} & \qgfjacval{59.3 \pm 8.9} & \qgfval{58.0 \pm 7.0} \\
         & T4  & \qgftrainval{0.0 \pm 0.0} & \qgftrainval{0.0 \pm 0.0} & \qgftrainval{15.0 \pm 7.7} & \qgftrainval{3.3 \pm 2.2} & \qgftrainval{65.3 \pm 6.1} & \qgftestval{5.0 \pm 3.7} & \qgftestval{4.7 \pm 3.2} & \qgftestval{38.3 \pm 7.3} & \qgftestval{36.0 \pm 7.6} & \qgftestval{6.3 \pm 3.6} & \qgfjacval{35.0 \pm 8.3} & \qgfval{52.0 \pm 11.6} \\
         & T5  & \qgftrainval{0.0 \pm 0.0} & \qgftrainval{0.0 \pm 0.0} & \qgftrainval{6.7 \pm 2.2} & \qgftrainval{29.3 \pm 12.5} & \qgftrainval{25.7 \pm 10.5} & \qgftestval{0.0 \pm 0.0} & \qgftestval{0.7 \pm 0.8} & \qgftestval{31.0 \pm 4.7} & \qgftestval{18.0 \pm 6.7} & \qgftestval{34.7 \pm 3.0} & \qgfjacval{31.3 \pm 10.0} & \qgfval{48.7 \pm 6.3} \\
         & Avg & \qgftrainval{0.8 \pm 0.6} & \qgftrainval{6.9 \pm 1.6} & \qgftrainval{31.2 \pm 5.0} & \qgftrainval{29.9 \pm 7.5} & \qgftrainval{74.3 \pm 2.7} & \qgftestval{18.5 \pm 1.8} & \qgftestval{34.9 \pm 3.0} & \qgftestval{60.7 \pm 4.5} & \qgftestval{50.3 \pm 4.5} & \qgftestval{37.1 \pm 3.0} & \qgfjacval{62.1 \pm 5.3} & \qgfval{70.1 \pm 4.7} \\
        \midrule
         & T1  & \qgftrainval{0.0 \pm 0.0} & \qgftrainval{0.7 \pm 0.9} & \qgftrainval{4.0 \pm 2.7} & \qgftrainval{16.0 \pm 9.6} & \qgftrainval{44.3 \pm 9.6} & \qgftestval{2.0 \pm 1.4} & \qgftestval{8.3 \pm 3.8} & \qgftestval{34.3 \pm 8.9} & \qgftestval{18.0 \pm 4.0} & \qgftestval{7.7 \pm 3.1} & \qgfjacval{44.0 \pm 9.3} & \qgfval{29.3 \pm 6.0} \\
         & T2  & \qgftrainval{0.0 \pm 0.0} & \qgftrainval{0.0 \pm 0.0} & \qgftrainval{0.0 \pm 0.0} & \qgftrainval{1.0 \pm 1.0} & \qgftrainval{19.7 \pm 8.2} & \qgftestval{0.0 \pm 0.0} & \qgftestval{3.3 \pm 2.6} & \qgftestval{6.7 \pm 3.5} & \qgftestval{1.7 \pm 1.0} & \qgftestval{0.3 \pm 0.6} & \qgfjacval{5.3 \pm 2.5} & \qgfval{5.3 \pm 4.0} \\
        \raisebox{-\baselineskip}[0pt][0pt]{\shortstack[l]{Cube\\Quadruple}} & T3  & \qgftrainval{0.0 \pm 0.0} & \qgftrainval{0.0 \pm 0.0} & \qgftrainval{0.0 \pm 0.0} & \qgftrainval{0.3 \pm 0.7} & \qgftrainval{5.0 \pm 2.8} & \qgftestval{0.0 \pm 0.0} & \qgftestval{0.7 \pm 0.8} & \qgftestval{0.0 \pm 0.0} & \qgftestval{0.0 \pm 0.0} & \qgftestval{0.0 \pm 0.0} & \qgfjacval{0.0 \pm 0.0} & \qgfval{0.0 \pm 0.0} \\
         & T4  & \qgftrainval{0.0 \pm 0.0} & \qgftrainval{0.0 \pm 0.0} & \qgftrainval{0.0 \pm 0.0} & \qgftrainval{0.0 \pm 0.0} & \qgftrainval{0.0 \pm 0.0} & \qgftestval{0.0 \pm 0.0} & \qgftestval{0.0 \pm 0.0} & \qgftestval{0.0 \pm 0.0} & \qgftestval{0.0 \pm 0.0} & \qgftestval{0.0 \pm 0.0} & \qgfjacval{0.0 \pm 0.0} & \qgfval{0.0 \pm 0.0} \\
         & T5  & \qgftrainval{0.0 \pm 0.0} & \qgftrainval{0.0 \pm 0.0} & \qgftrainval{0.0 \pm 0.0} & \qgftrainval{0.0 \pm 0.0} & \qgftrainval{0.0 \pm 0.0} & \qgftestval{0.0 \pm 0.0} & \qgftestval{0.0 \pm 0.0} & \qgftestval{0.0 \pm 0.0} & \qgftestval{0.0 \pm 0.0} & \qgftestval{0.0 \pm 0.0} & \qgfjacval{0.0 \pm 0.0} & \qgfval{0.0 \pm 0.0} \\
         & Avg & \qgftrainval{0.0 \pm 0.0} & \qgftrainval{0.1 \pm 0.2} & \qgftrainval{0.8 \pm 0.5} & \qgftrainval{3.5 \pm 1.9} & \qgftrainval{13.8 \pm 3.3} & \qgftestval{0.4 \pm 0.3} & \qgftestval{2.5 \pm 0.8} & \qgftestval{8.2 \pm 2.2} & \qgftestval{3.9 \pm 0.9} & \qgftestval{1.6 \pm 0.7} & \qgfjacval{9.9 \pm 2.3} & \qgfval{6.9 \pm 1.8} \\
        \midrule
         & T1  & \qgftrainval{1.0 \pm 1.0} & \qgftrainval{0.0 \pm 0.0} & \qgftrainval{41.0 \pm 9.5} & \qgftrainval{57.3 \pm 7.8} & \qgftrainval{65.0 \pm 4.3} & \qgftestval{23.0 \pm 5.9} & \qgftestval{47.3 \pm 5.0} & \qgftestval{25.3 \pm 5.3} & \qgftestval{49.3 \pm 4.1} & \qgftestval{58.7 \pm 6.0} & \qgfjacval{40.0 \pm 4.0} & \qgfval{57.7 \pm 5.2} \\
         & T2  & \qgftrainval{0.3 \pm 0.7} & \qgftrainval{0.0 \pm 0.0} & \qgftrainval{0.0 \pm 0.0} & \qgftrainval{22.7 \pm 3.0} & \qgftrainval{16.3 \pm 4.9} & \qgftestval{0.0 \pm 0.0} & \qgftestval{30.7 \pm 6.8} & \qgftestval{14.3 \pm 4.9} & \qgftestval{26.0 \pm 6.6} & \qgftestval{34.7 \pm 5.6} & \qgfjacval{34.7 \pm 5.9} & \qgfval{34.7 \pm 5.3} \\
        \raisebox{-0.5\baselineskip}[0pt][0pt]{Puzzle $4\times4$} & T3  & \qgftrainval{0.3 \pm 0.7} & \qgftrainval{0.0 \pm 0.0} & \qgftrainval{36.7 \pm 16.3} & \qgftrainval{67.0 \pm 3.7} & \qgftrainval{73.0 \pm 6.6} & \qgftestval{12.0 \pm 4.5} & \qgftestval{44.3 \pm 8.4} & \qgftestval{27.3 \pm 5.3} & \qgftestval{44.7 \pm 5.5} & \qgftestval{64.3 \pm 4.3} & \qgfjacval{43.3 \pm 5.3} & \qgfval{65.3 \pm 3.9} \\
         & T4  & \qgftrainval{0.3 \pm 0.7} & \qgftrainval{0.0 \pm 0.0} & \qgftrainval{4.7 \pm 4.8} & \qgftrainval{27.0 \pm 6.7} & \qgftrainval{22.7 \pm 5.1} & \qgftestval{0.0 \pm 0.0} & \qgftestval{26.3 \pm 5.8} & \qgftestval{14.3 \pm 2.8} & \qgftestval{26.0 \pm 2.9} & \qgftestval{32.7 \pm 5.1} & \qgfjacval{24.7 \pm 4.5} & \qgfval{42.0 \pm 5.7} \\
         & T5  & \qgftrainval{1.0 \pm 1.0} & \qgftrainval{0.0 \pm 0.0} & \qgftrainval{0.0 \pm 0.0} & \qgftrainval{24.0 \pm 8.2} & \qgftrainval{14.0 \pm 5.6} & \qgftestval{1.0 \pm 1.0} & \qgftestval{14.0 \pm 4.6} & \qgftestval{10.3 \pm 2.5} & \qgftestval{23.3 \pm 7.0} & \qgftestval{31.7 \pm 4.3} & \qgfjacval{22.7 \pm 4.3} & \qgfval{38.3 \pm 7.1} \\
         & Avg & \qgftrainval{0.6 \pm 0.3} & \qgftrainval{0.0 \pm 0.0} & \qgftrainval{16.5 \pm 4.7} & \qgftrainval{39.6 \pm 3.2} & \qgftrainval{38.2 \pm 2.8} & \qgftestval{7.2 \pm 1.0} & \qgftestval{32.5 \pm 1.8} & \qgftestval{18.3 \pm 2.6} & \qgftestval{33.9 \pm 1.8} & \qgftestval{44.4 \pm 2.8} & \qgfjacval{33.1 \pm 2.2} & \qgfval{47.6 \pm 2.5} \\
        \midrule
         & T1  & \qgftrainval{39.3 \pm 3.6} & \qgftrainval{76.3 \pm 6.1} & \qgftrainval{100.0 \pm 0.0} & \qgftrainval{100.0 \pm 0.0} & \qgftrainval{100.0 \pm 0.0} & \qgftestval{92.7 \pm 7.4} & \qgftestval{99.3 \pm 0.8} & \qgftestval{94.7 \pm 2.1} & \qgftestval{100.0 \pm 0.0} & \qgftestval{100.0 \pm 0.0} & \qgfjacval{100.0 \pm 0.0} & \qgfval{99.7 \pm 0.6} \\
         & T2  & \qgftrainval{8.7 \pm 3.8} & \qgftrainval{38.7 \pm 8.8} & \qgftrainval{99.3 \pm 0.9} & \qgftrainval{99.7 \pm 0.7} & \qgftrainval{100.0 \pm 0.0} & \qgftestval{5.3 \pm 5.0} & \qgftestval{99.0 \pm 1.9} & \qgftestval{81.0 \pm 3.5} & \qgftestval{100.0 \pm 0.0} & \qgftestval{97.3 \pm 1.5} & \qgfjacval{97.7 \pm 1.3} & \qgfval{92.7 \pm 3.3} \\
        \raisebox{-0.5\baselineskip}[0pt][0pt]{Scene} & T3  & \qgftrainval{14.3 \pm 5.0} & \qgftrainval{20.0 \pm 6.6} & \qgftrainval{73.0 \pm 12.2} & \qgftrainval{100.0 \pm 0.0} & \qgftrainval{98.7 \pm 1.4} & \qgftestval{0.0 \pm 0.0} & \qgftestval{99.7 \pm 0.6} & \qgftestval{89.7 \pm 2.8} & \qgftestval{99.7 \pm 0.6} & \qgftestval{99.3 \pm 0.8} & \qgfjacval{99.3 \pm 0.8} & \qgfval{97.0 \pm 1.1} \\
         & T4  & \qgftrainval{17.7 \pm 5.0} & \qgftrainval{39.7 \pm 7.4} & \qgftrainval{71.7 \pm 12.8} & \qgftrainval{71.3 \pm 9.1} & \qgftrainval{54.0 \pm 21.2} & \qgftestval{6.3 \pm 3.7} & \qgftestval{89.3 \pm 11.0} & \qgftestval{55.7 \pm 9.2} & \qgftestval{71.0 \pm 8.5} & \qgftestval{80.0 \pm 4.4} & \qgfjacval{79.0 \pm 6.2} & \qgfval{85.3 \pm 5.2} \\
         & T5  & \qgftrainval{4.7 \pm 3.1} & \qgftrainval{4.0 \pm 1.9} & \qgftrainval{26.3 \pm 12.4} & \qgftrainval{77.0 \pm 10.0} & \qgftrainval{37.7 \pm 13.6} & \qgftestval{0.0 \pm 0.0} & \qgftestval{97.7 \pm 2.1} & \qgftestval{20.7 \pm 4.1} & \qgftestval{64.0 \pm 5.4} & \qgftestval{52.7 \pm 4.9} & \qgfjacval{66.0 \pm 5.0} & \qgfval{60.3 \pm 4.5} \\
         & Avg & \qgftrainval{16.9 \pm 1.6} & \qgftrainval{35.7 \pm 2.4} & \qgftrainval{74.1 \pm 4.3} & \qgftrainval{89.6 \pm 2.6} & \qgftrainval{78.1 \pm 3.6} & \qgftestval{20.9 \pm 2.2} & \qgftestval{97.0 \pm 2.5} & \qgftestval{68.3 \pm 2.3} & \qgftestval{86.9 \pm 2.5} & \qgftestval{85.9 \pm 1.7} & \qgfjacval{88.4 \pm 1.3} & \qgfval{87.0 \pm 2.0} \\
        \bottomrule
    \end{tabular}
    }%
    \caption{Numeric details of \Cref{fig:full-result-no-bfn}. Results report mean with $95\%$ confidence interval across $10$ random seeds.}
    \label{tab:main-results}
\end{table}

\begin{figure}[h]
    \centering
    \includegraphics[width=\linewidth]{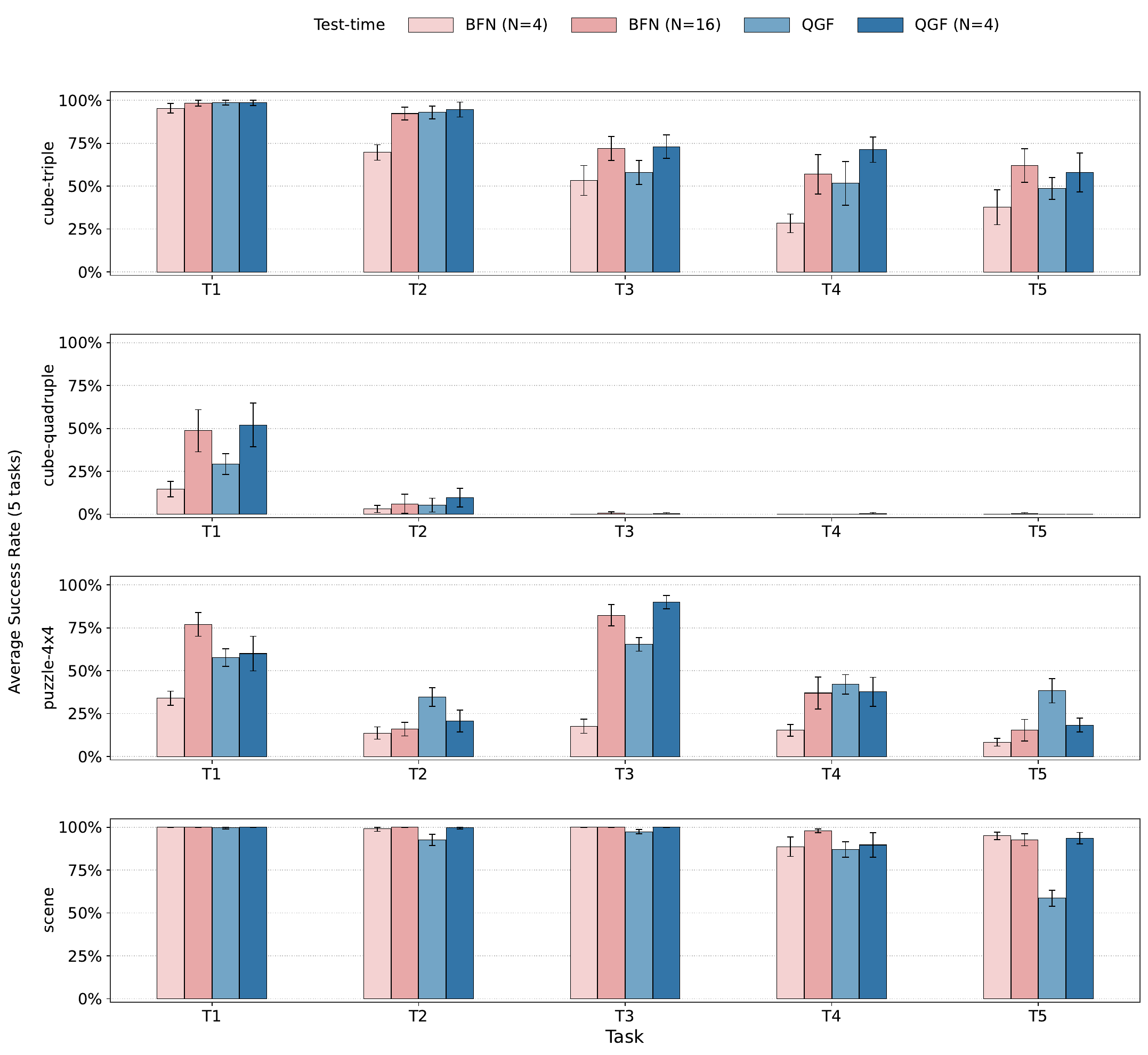}
    \vspace{-0.5cm}
    \caption{Task breakdown of \Cref{fig:result-bfn}: offline RL with Best-of-N sampling at 500k training steps per environment ($10$ seeds each).}
    \label{fig:full-result-bfn}
\end{figure}

\newpage 
\begin{figure}[h]
    \centering
    \includegraphics[width=0.9\linewidth]{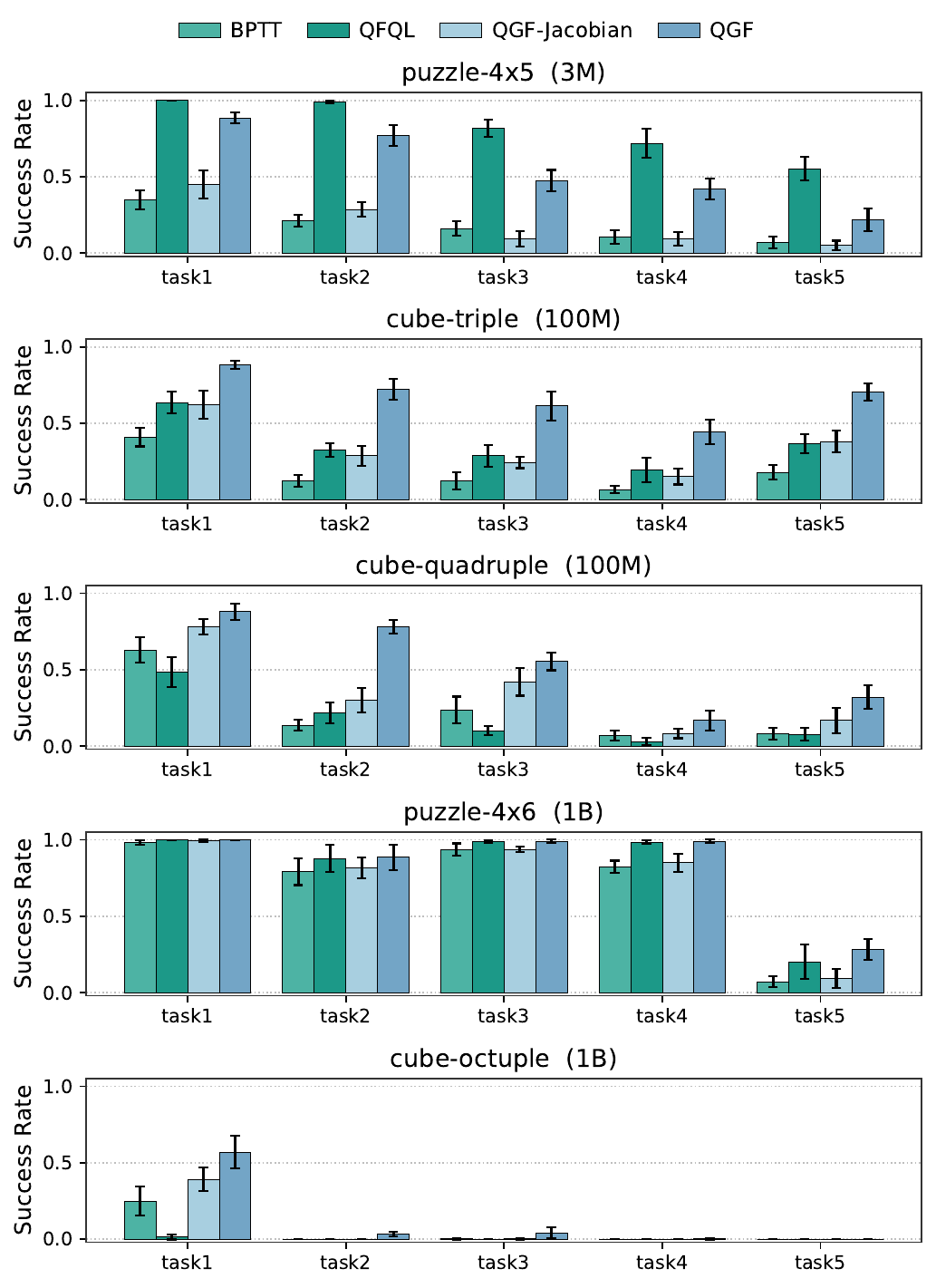}
    \caption{Task breakdown of \Cref{fig:dqc-results}: offline goal-conditioned RL at 1M training steps (10 seeds each). See the domain-specific hyperparameters in \cref{tab:domain_hyperparams_dqc}.}
    \label{fig:full-result-dqc}
\end{figure}

\begin{figure}[h]
    \centering
    \includegraphics[width=\linewidth]{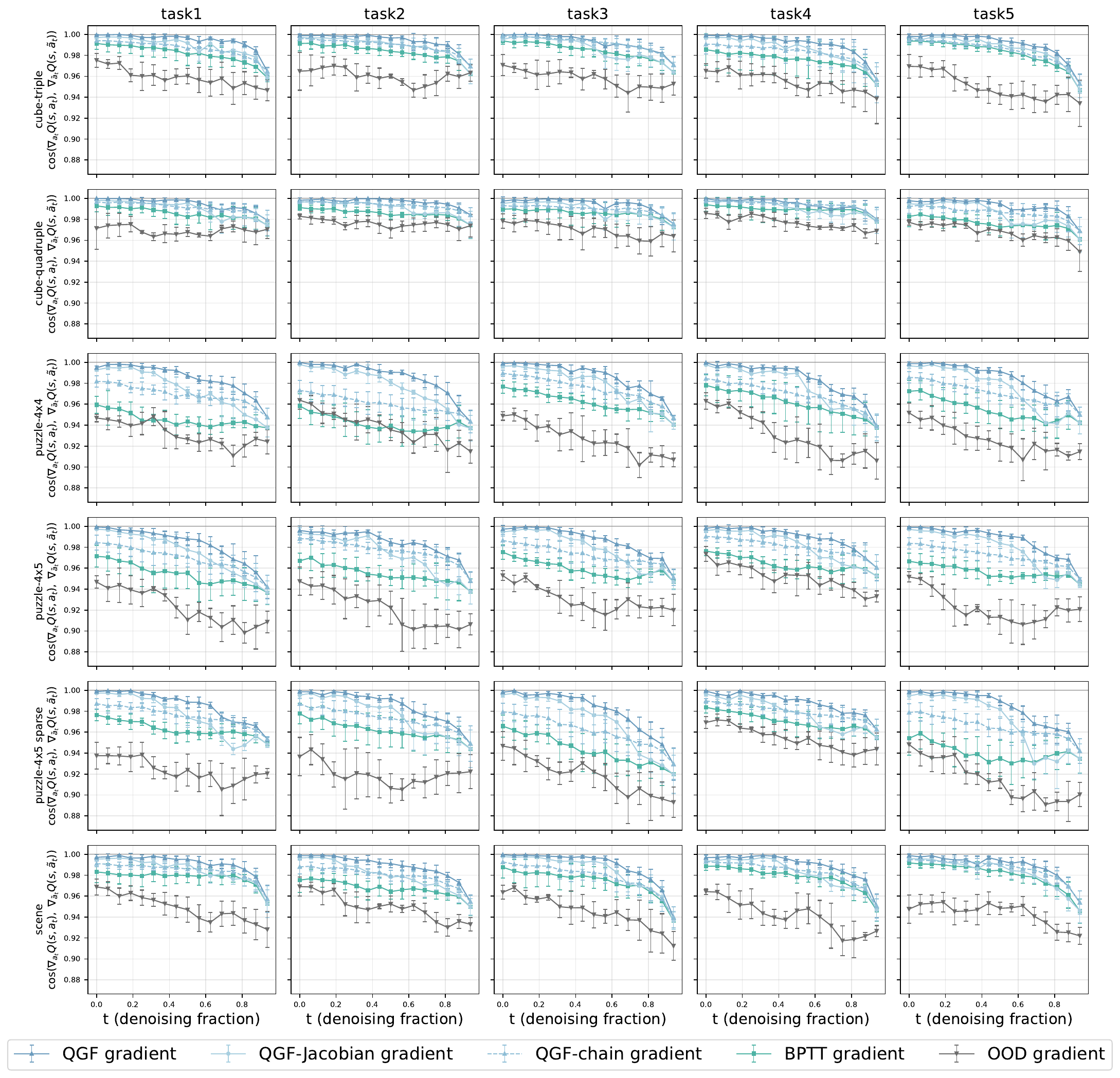}
    \caption{Task breakdown of \Cref{fig:gradient-estimator-variance}: on sensitivity of gradient estimator to noise for all $20$ OGBench environments. $4$ seeds each.}
    \label{fig:gradient-estimator-variance-full-results}
\end{figure}

\end{document}